\pdfoutput=1

\documentclass[11pt]{article}

\usepackage[final]{acl}

\def\code#1{\texttt{#1}}
\usepackage{xspace}

\usepackage{array, boldline, makecell, booktabs}
\usepackage[svgnames, table]{xcolor}

\makeatletter
\NewDocumentCommand{\supptitle}{s}{
\twocolumn[{
  \begin{center}
    \vspace*{-0.3cm}
    \rule{\textwidth}{0.05cm}\\[0.1cm]
    \textbf{- Appendix -}\\[0.2cm]
    {\Large \textbf{\mytitle}}\\[0.1cm]
    \rule{\textwidth}{0.05cm}\\[0.3cm]
  \end{center}
}]
}
\makeatother

\newcommand{\cmark}{\textcolor{black}{\ding{51}}\xspace}%
\newcommand{\xmark}{\textcolor{black}{\ding{55}}\xspace}%

\definecolor{LightCyan}{rgb}{0.88,1,1}
\definecolor{Blue}{rgb}{0, 0.5, 1}
\definecolor{Green}{rgb}{0.0, 0.8, 0.0 }
\definecolor{Red}{rgb}{0.95, 0.55, 0.6}
\definecolor{Skyblue}{rgb}{0.6, 0.6, 0.95 }
\definecolor{Beige}{rgb}{0.96, 0.96, 0.86}

\newcommand{\alg}{\code{PPA-Plan}\xspace}
\newcommand{\mytitle}{\alg : Proactive Pitfall Avoidance for Reliable Planning in Long-Context LLM Reasoning}

\usepackage[normalem]{ulem}
\useunder{\uline}{\ul}{}

\usepackage{tcolorbox}
\tcbuselibrary{skins, breakable}

\usepackage[hang,flushmargin]{footmisc}
\definecolor{softblue}{RGB}{240, 245, 255}

\usepackage{subcaption}

\usepackage{pifont}

\usepackage{amsmath,amsfonts,bm}

\def\eqref#1{equation~\ref{#1}}

\def\1{\bm{1}}

\DeclareMathAlphabet{\mathsfit}{\encodingdefault}{\sfdefault}{m}{sl}
\SetMathAlphabet{\mathsfit}{bold}{\encodingdefault}{\sfdefault}{bx}{n}

\usepackage{times}
\usepackage{latexsym}
\usepackage{tikz}
\usetikzlibrary{positioning, arrows.meta, shapes.geometric, calc, decorations.pathreplacing}

\usepackage{amsmath}  
\usepackage{amssymb}  
\usepackage{xcolor}  
\usepackage[T1]{fontenc}

\usepackage[utf8]{inputenc}

\usepackage{microtype}

\usepackage{inconsolata}

\usepackage{graphicx}

\usepackage{capt-of}

\usepackage{algorithm}
\usepackage{footmisc}

\usepackage{newfloat}
\usepackage{listings}
\DeclareCaptionStyle{ruled}{labelfont=normalfont,labelsep=colon,strut=off} 
\floatstyle{ruled}
\newfloat{listing}{tb}{lst}{}
\floatname{listing}{Listing}
\usepackage{times}
\usepackage{latexsym}
\usepackage{mathtools}
\usepackage{amsthm}

\usepackage[capitalize,noabbrev]{cleveref}

\usepackage{tikz}
\usetikzlibrary{fadings}
\usetikzlibrary{patterns}
\usetikzlibrary{shadows.blur}
\usetikzlibrary{shapes}

\usepackage{latexsym}
\usepackage{nccmath}
\usepackage{bbm}
\usepackage{tabularx}
\usepackage{yhmath}
\usepackage{times}
\usepackage{latexsym}
\usepackage{booktabs}
\usepackage{graphicx}
\usepackage{algorithm}
\usepackage{subcaption}
\usepackage{algpseudocode}
\usepackage{amsmath}
\usepackage{siunitx}
\usepackage{mathtools}
\usepackage{array}
\usepackage{multirow}
\usepackage[normalem]{ulem}
\usepackage{arydshln}

\useunder{\uline}{\ul}{}
 
\makeatletter
\newcommand{\multiline}[1]{%
  \begin{tabularx}{\dimexpr\linewidth-\ALG@thistlm}[t]{@{}X@{}}
    #1
  \end{tabularx}
}
\makeatother

\usepackage[capitalize]{cleveref}
\crefname{section}{Sec.}{Secs.}
\Crefname{section}{Section}{Sections}
\Crefname{table}{Table}{Tables}
\crefname{table}{Tab.}{Tabs.}

\algblock{Input}{EndInput}
\algnotext{EndInput}
\algblock{Output}{EndOutput}
\algnotext{EndOutput}

\newif\ifshowcomments
\showcommentsfalse

\title{\mytitle}

\author{
  \textbf{Byeongjin Kim}$^{1}$ \quad \textbf{Gyuwan Kim}$^{2}$ \quad \textbf{Seo Yeon Park}$^{1\dagger}$ \\
  $^1$Hanyang University \quad $^2$University of California, Santa Barbara \\
  \texttt{\{qudwls5828, seoyeonpark\}@hanyang.ac.kr}, \texttt{gyuwankim@ucsb.edu}
}

\begin{document}

\maketitle
\def\thefootnote{$\dagger$}\footnotetext{Corresponding author}
\setcounter{footnote}{0}
\enlargethispage{1\baselineskip}

\begin{abstract}

Large language models struggle with reasoning over long contexts where relevant information is sparsely distributed. Although plan-and-execute frameworks mitigate this by decomposing tasks into planning and execution, their effectiveness is often limited by unreliable plan generation due to dependence on surface-level cues.
Consequently, plans may be based on incorrect assumptions, and once a plan is formed, identifying errors and revising it reliably becomes difficult, limiting the effectiveness of reactive refinement.
To address this limitation, we propose \alg{}, a proactive planning strategy for long-context reasoning that focuses on preventing such failures before plan generation.
\alg{} identifies potential logical pitfalls and false assumptions, formulates them as negative constraints, and conditions plan generation on explicitly avoiding these constraints. Experiments on long-context QA benchmarks show that executing plans generated by \alg consistently outperforms existing plan-and-execute methods and direct prompting.
\end{abstract}

\section{Introduction}

\begin{figure*}[t]
\centering
\includegraphics[width=1.0\textwidth]{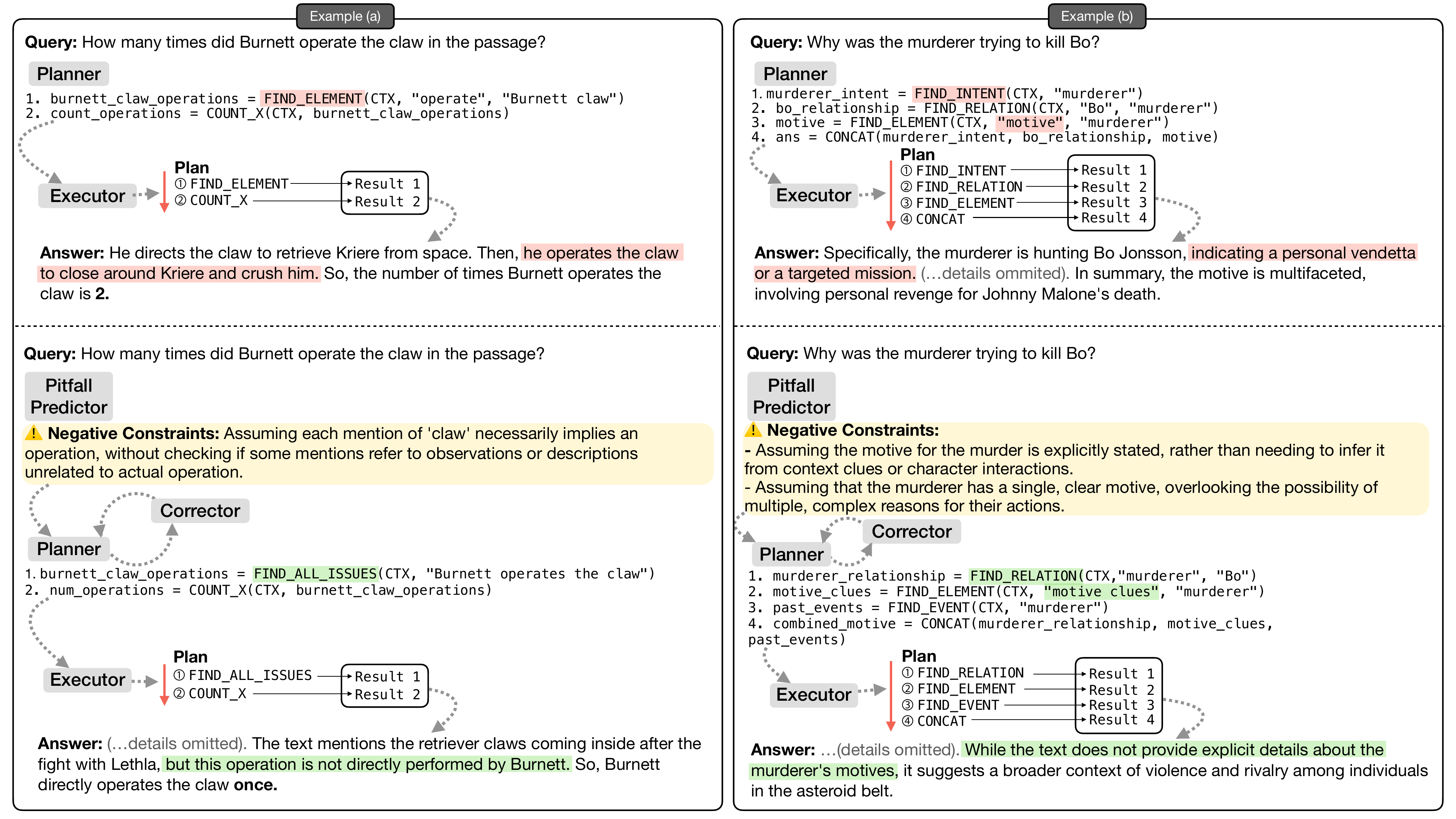}
\caption{
Comparison of planning and execution results by PEARL~\citep{sun-etal-2024-pearl} (top) and our \alg{} (bottom) on QuALITY~\citep{pang-etal-2022-quality} using GPT-4o-mini. 
In Example (a), PEARL produces an incorrect answer to a counting query due to surface-level keyword matching that ignores the situational context of who performed each action.
In Example (b), PEARL hallucinates a motive for a murder despite the absence of supporting evidence in the document, resulting from premature assumptions.
In contrast, \alg{} explicitly identifies and avoids such logical pitfalls during plan generation, leading to more reliable reasoning and answers.
}
\label{fig:introduction_example}
\end{figure*} 
Large language models (LLMs) have become increasingly strong~\citep{brown2020language, srivastava2023beyond}, with substantial improvements in both reasoning performance and supported context length, enabling their use across a wide range of applications.
Nevertheless, when moving beyond surface-level information retrieval or local understanding to complex, multi-step reasoning over long inputs, significant challenges remain.
In such settings, task-relevant evidence is often sparsely distributed across distant spans, mixed with irrelevant details, or presented in positions that bias model attention~\citep{liu-etal-2024-lost}. This leads to failure modes such as missed evidence, unsupported assumptions, hallucination, and incorrect answers.
Even with standard chain-of-thought prompting~\citep{wei2022chain}, models frequently struggle to construct coherent reasoning trajectories and fail to arrive at correct answers for complex queries, particularly in long-context scenarios.
These observations suggest that effective long-context reasoning requires not only more extensive reasoning traces, but also more deliberate control over how reasoning processes are structured and guided.

To address these challenges, recent work has proposed plan-and-execute frameworks that decompose complex long-context tasks into an explicit planning stage followed by step-by-step execution~\citep{yao2023react, wang-etal-2023-plan, sun-etal-2024-pearl, erdoganplan, hu2025hiagent}. 
While these approaches often leverage LLMs to generate intermediate plans and have demonstrated improvements over direct generation, the planning stage itself remains a critical bottleneck. This is because 
existing LLM-based planners typically rely on surface-level textual patterns present in long-context inputs when generating plans (as illustrated in~\autoref{fig:introduction_example}), rather than faithfully capturing the underlying logical structure of the instruction or task~\citep{valmeekam2023planbench, tang-etal-2023-large}. 
As a result, generated plans often embed implicit or unsupported assumptions, overlook necessary constraints, or commit prematurely to specific reasoning paths, causing errors to cascade throughout subsequent execution.

To mitigate such issues, prior work has largely adopted reactive refinement strategies that revise plans~\cite{sun-etal-2024-pearl}.
However, this approach is fundamentally limited, as LLMs tend to anchor on their own generated outputs and exhibit strong resistance to revising incorrect assumptions once they are formed~\cite{huang2023large, xi2025rise}. 
Consequently, identifying and correcting errors after plan generation is often more difficult than preventing them in advance at the planning stage.

To equip LLMs with the ability to generate reliable plans to execute them for long-context scenarios, we introduce a planning strategy, \alg, short for \textbf{\uline{P}}roactive \textbf{\uline{P}}itfall \textbf{\uline{A}}voidance \textbf{\uline{Plan}}ning.
\alg{} consists of three modules: (1) a \textit{Pitfall Predictor}, (2) a \textit{Planner}, and (3) a \textit{Corrector}. 
First, the \textit{Pitfall Predictor} analyzes a query to explore potential logical pitfalls and false assumptions, which are formulated as \textit{Negative Constraints} that should be avoided during the plan generation (see \autoref{fig:introduction_example}).
Conditioned on these constraints, the \textit{Planner} performs \textit{Strategy Reasoning} to determine the logical structure of a plan while satisfying all the specified negative constraints.
After this, if the plan violates syntactic requirements for executability, the \textit{Corrector} is iteratively invoked to repair such errors via \textit{Strategy Reasoning}, while preserving the original intent by referencing both the query and the negative constraint set to preserve the initial plan logic. 
Once a valid plan is obtained, we sequentially execute its steps to produce an open-ended free-form output using LLMs. 
From a broader perspective, \alg{} can also be viewed as a modular, multi-agent system, where different components specialize in distinct reasoning roles and interact through explicit intermediate outputs~\citep{woolridge2001introduction, yao2023react, wu2024autogen, kim2024llm}.

We evaluate the effectiveness of \alg{} on a subset of long-context question-answering benchmarks, including QuALITY~\citep{pang-etal-2022-quality}, ConditionalQA~\citep{sun-etal-2022-conditionalqa}, and LongReason~\citep{ling2025longreason}, and Qasper \citep{dasigi-etal-2021-dataset}, a reading comprehension dataset that contains questions about long-form articles. 
For multiple-choice question-answering datasets, we map the model's generated free-form answer to one of the candidate options using an LLM and report accuracy, following~\citet{sun-etal-2024-pearl}.
Executing plans generated by \alg{} consistently yields more accurate answers with improved coverage than existing plan-and-execute strategies or direct prompting baselines, particularly for questions that require reasoning over the full long document.

\section{Related Work}

\paragraph{Long-Context Reasoning.}
Recent advancements in LLM architectures and training systems, such as Rotary Position Embedding (RoPE)~\cite{su2024roformer}, Position Interpolation (PI)~\cite{chen2023extending}, RingAttention~\cite{liu2023ring}, YaRN~\cite{peng2024yarn}, and ALiBi~\cite{peng2024yarn} have significantly enhanced the model’s capacity to process extended context lengths; however, the ability to perform complex reasoning over long contexts remains limited, as expanding the context window does not inherently grant the model the capability to synthesize logic across vast information~\cite{liu-etal-2024-lost}. This limitation, further evidenced by comprehensive long-context benchmarks like LongBench~\cite{bai-etal-2024-longbench}, LooGLE~\cite{li2024loogle}, and RULER~\cite{hsieh2024rulerwhatsrealcontext}, as well as reasoning-intensive tasks in QuALITY~\cite{pang-etal-2022-quality} and LongReason~\cite{ling2025longreason}, motivates our work to focus on enhancing reasoning strategies rather than merely extending input length.

\paragraph{LLM-based Planning.}
Foundational planning frameworks, such as Plan-and-Solve~\citep{wang-etal-2023-plan} and ReAct~\citep{yao2023react}, established the importance of task decomposition; however, they are not inherently designed for long-context scenarios. Subsequent reactive strategies like Reflexion~\citep{shinn2023reflexion}, Self-Refine~\citep{madaan2023self}, and ADaPT~\citep{prasad2024adapt} utilize linguistic feedback to refine logic post-hoc, yet they tend to anchor on initial faulty outputs~\citep{huang2023large, xi2025rise}. While PEARL~\cite{sun-etal-2024-pearl} successfully integrated these paradigms into long-context tasks, it remains limited by its reliance on reactive refinement and a mechanism designed around surface-level textual patterns. These limitations motivate our focus on proactively preventing logical errors by leveraging an agent-planning framework designed for long-context reasoning.

\begin{figure*}[t]
\centering
\includegraphics[width=1.0\textwidth]{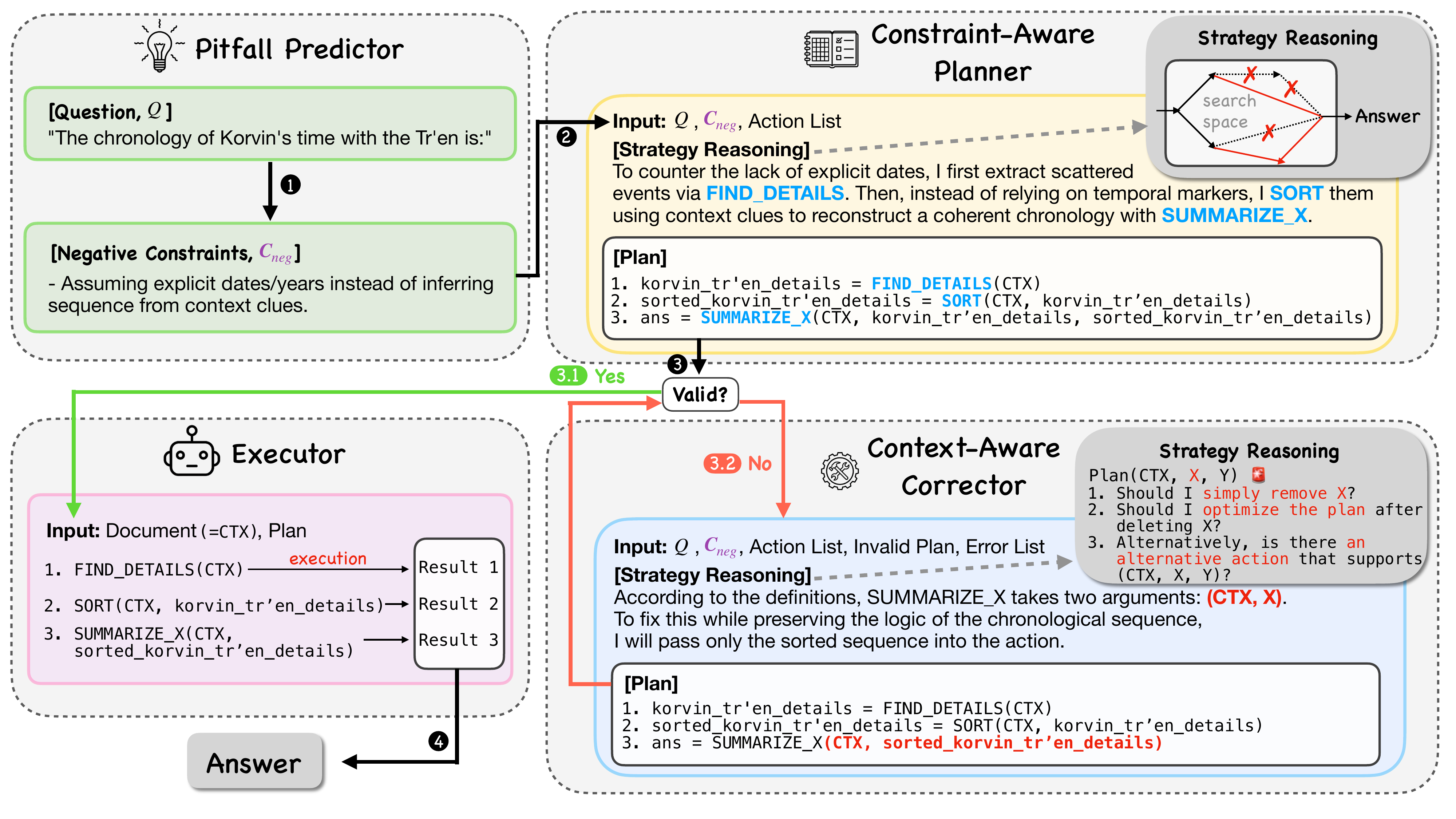}
\caption{
Overview of \alg, a proactive planning framework designed to generate reliable plans and execute them for long-context reasoning.
The figure illustrates the full planning process through a concrete example.
(1) If the document is not expected to contain explicit temporal markers based on the query, $\mathcal{M}_{pred}$ generates negative constraints to suppress the assumption of concrete dates.
(2) Guided by these constraints, $\mathcal{M}_{plan}$ performs strategy reasoning to reconstruct the chronology from scattered events based on context clues, rather than attempting a futile search for nonexistent explicit dates, then $\mathcal{M}_{plan}$ generates a plan based on this strategy.
(3) If the plan has an invalid format, such as incorrect arguments of \texttt{SUMMARIZE\_X}, $\mathcal{M}_{corr}$ conducts strategy analysis to find the best way of fixing and optimizing the plan, such as by removing redundant arguments.
}
\label{fig:main_figure}
\end{figure*}

\section{Method}
We propose a planning strategy, \alg{}, that produces reliable execution plans for tasks that require complex reasoning over long documents.
The core idea of \alg{} is to proactively identify potential logical pitfalls and false assumptions before plan generation, and to generate negative constraints that explicitly guide the planning process.
By conditioning planning on these constraints, \alg{} systematically avoids error-prone reasoning paths that are difficult to correct once a plan is formed.
\alg consists of three components: (1) a pitfall predictor $\mathcal{M}_{pred}$, (2) a planner $\mathcal{M}_{plan}$, and (3) a corrector $\mathcal{M}_{corr}$. 
\autoref{fig:main_figure} illustrates the overall framework of \alg{} and a concrete example.

\subsection{Pitfall Predictor for Negative Constraint Generation}
\label{sec:1}

The \textit{pitfall predictor} $\mathcal{M}_{pred}$ analyzes a query $q$ to identify potential logical pitfalls and false assumptions, which are formalized as a negative constraint set $\mathcal{C}_{neg}$.
These constraints specify reasoning patterns that should be avoided during plan generation.
To perform this analysis, $\mathcal{M}_{pred}$ adopts the functional roles of an \emph{exam designer} and a \emph{logic analyst}, critically examining $q$ to uncover implicit premises.
Using a structured prompt, as shown in \autoref{tab:neg_const}, with task-specific guidelines and few-shot demonstrations covering multi-hop inference, scope constraints, and counting, the predictor identifies risks such as shallow semantic patterns, scope confusion, and keyword-based guessing.
The predictor derives up to $k$ most critical constraints in a structure JSON format to ensure compatibility with the downstream modules. We formalize this process as follows:
\[
    \mathcal{C}_{neg} = \{c_1, c_2, \dots, c_k\} = \mathcal{M}_{pred}(q)
\]

\noindent The resulting $\mathcal{C}_{neg}$ is passed to the planner to facilitate constraint-aware plan generation.

\subsection{Constraint-Aware Plan Generation}
\label{sec:2}

Given the query $q$ and negative constraint set $\mathcal{C}_{neg}$, the \textit{planner} $\mathcal{M}_{plan}$ generates an executable plan while avoiding the identified logical pitfalls.
Rather than directly producing a plan, $\mathcal{M}_{plan}$ first performs \textit{Strategy Reasoning}, an intermediate reasoning step that determines how the action sequence for the query $q$ should be structured to satisfy the constraints, by analyzing $\mathcal{C}_{neg}$.

This reasoning step is guided by few-shot demonstrations that map the identified pitfalls to concrete strategic adjustments (\autoref{tab:CA-Plan}).
By incorporating exemplars that showcase the causal link between strategy and action selection, the module guides the agent's internal reasoning process toward the intended cognitive path before generating the initial plan.
The rationale behind this approach (Strategy Reasoning) is that simply imposing $\mathcal{C}_{neg}$ often leads models to ignore prohibitions or leaves them unsure of alternative actions.

To ensure planning stability and reliable grounding, $\mathcal{M}_{plan}$ selects actions from a pre-defined action space $\mathcal{A}$\footnote{
We adopt the standard action space $\mathcal{A}$ and function-call format \texttt{output=action(args)} provided by PEARL~\citep{sun-etal-2024-pearl}. We also allow $\mathcal{M}_{plan}$ to dynamically define and use auxiliary actions if the required operation is absent from $\mathcal{A}$.} to translate this derived strategy into an executable plan. 
This approach prevents the generation of non-executable text, aligning the agent's logic with the environment's specific capabilities.
Since all necessary evidence is already contained in the provided context, designing an optimized trajectory at the pre-execution stage is sufficient for long-context QA tasks. This avoids the additional computational overhead of multi-round approaches and improves efficiency and scalability by reducing the number of model calls. 
The initial plan is generated as follows:
\[
    \mathcal{P}^{(0)} = \mathcal{M}_{plan}(q, \mathcal{A}, \mathcal{C}_{neg})
\]

\noindent Once $\mathcal{P}^{(0)}$ is generated, a syntactic validity check is applied.
If the plan is executable, it is forwarded to the execution module. 
Otherwise, it is passed to the corrector for refinement.

\subsection{Context-Aware Correction}
\label{sec:3}

Simultaneously managing high-level logic and strict formatting often leads LLMs to prioritize semantic integrity over syntactic precision \cite{saparov2023language, zhao2025trade}. To resolve these issues, we propose \textit{Context-Aware Correction}, which decouples syntactic alignment from logical planning. 
The corrector $\mathcal{M}_{corr}$ receives $q$ and $\mathcal{C}_{neg}$ alongside immediate feedback $\mathcal{F}^{(t-1)}$, defined as the invalid plan and associated error messages from the previous step ($t-1$). 
By focusing on $\mathcal{F}^{(t-1)}$ over a history, we mitigate the risk of informational noise from stale iterations and prioritize diagnosing and repairing the current failure.

$\mathcal{M}_{corr}$ performs Strategy Reasoning by mapping the error feedback $\mathcal{F}^{(t-1)}$ to a repair operation to rectify the flawed plan $\mathcal{P}^{(t-1)}$ generated for $q$.
$\mathcal{M}_{corr}$ accordingly determines the optimal resolution that satisfies syntactic requirements without compromising the original intent.
Based on this derived strategy, $\mathcal{M}_{corr}$ then reconstructs the invalid plan $\mathcal{P}^{(t-1)}$ into a refined version to achieve syntactic alignment.

To implement this, we employ few-shot in-context learning with demonstrations that cover representative failure cases, such as unknown actions, undefined variables, and incorrect argument counts (see \autoref{tab:CA-Corr}).
By showcasing the causal link between specific error types and their optimal resolutions, these few-shot demonstrations facilitate the strategy reasoning required to transform raw error messages into precise, executable plan corrections.
We formalize the iterative correction process as follows:
\[
    \mathcal{P}^{(t)} = \mathcal{M}_{corr}(q, \mathcal{A}, \mathcal{C}_{neg}, \mathcal{F}^{(t-1)}), \; 1 \leq t \leq B
\]

\noindent This process repeats until a plan satisfies the syntax requirements for execution or reaches the budget on the number of corrections, $B$.

\subsection{Plan Execution}
Once the final executable plan is obtained, it is passed to the plan execution module $\mathcal{M}_{exec}$.
The executor operates by sequentially parsing each step of the plan. 
For each step, it constructs a specific execution prompt by integrating the defined action, arguments, and the long document $\mathcal{D}$.
The resulting output from this execution is stored in a designated variable. 
Crucially, for steps requiring inputs from preceding actions, the executor resolves data dependencies by substituting variable placeholders with their corresponding stored results. 
This iterative process converts the logical plan into a set of concrete evidence $\mathcal{E}$, which constitutes the final system response.

\section{Experiments}
\subsection{Experimental Setup}

\paragraph{Model Selection.} We evaluate performance different LLMs: GPT-4o-mini~\cite{openai2024gpt4ocard}, Llama-3.1-8B-Instruct~\cite{grattafiori2024llama3herdmodels}, and Qwen-2.5-14B-Instruct~\cite{qwen2025qwen25technicalreport}.
For Llama and Qwen models, we apply 8-bit quantization~\cite{dettmers2022llmint88bitmatrixmultiplication} to maximize inference efficiency while minimizing performance loss. This approach ensures optimal performance within limited VRAM resources.
Specific hyperparameter settings and hardware specifications are described in the Appendix~\ref{sec:appendix_exp_details}.


\paragraph{Datasets and Task Formulation.} To verify complex reasoning in long contexts, we use QuALITY~\cite{pang-etal-2022-quality}, ConditionalQA~\cite{sun-etal-2022-conditionalqa}, and LongReason~\cite{ling2025longreason} for multiple-choice tasks and Qasper~\cite{dasigi-etal-2021-dataset} for free-form answer tasks.
We simulate real-world scenarios by restricting models from accessing gold-standard information; instead, we convert all instances into generative tasks that require long-form answers.
For \autoref{tab:main_results}, we evaluate methods on the intersection of samples for which all methods successfully generated outputs to ensure a fair qualitative comparison, while \autoref{tab:ablation} includes the full test set to better evaluate overall robustness.
For input length $L$ exceeding the context length $M$, we truncate the input sequence $S$ to preserve the document's extremities, as critical information is typically concentrated at the beginning and end~\citep{liu-etal-2024-lost}: $S_{1:L} \rightarrow [S_{1:\lfloor M/2 \rfloor} ; S_{L-\lfloor M/2 \rfloor - 1:L}]$~\citep{bai2024longbench}.
Detailed dataset statistics are provided in the Appendix~\ref{sec:appendix_exp_details}.

\definecolor{softblue}{RGB}{240, 245, 255}
\begin{table*}[t]
\centering

\setlength{\tabcolsep}{4pt}
\resizebox{0.9\textwidth}{!}{
\begin{tabular}{l c ccc cc ccc cc ccc}
\toprule
& & \multicolumn{3}{c}{\textbf{QuALITY}} & \multicolumn{2}{c}{\textbf{Cond.QA}} & \multicolumn{3}{c}{\textbf{LongReason}} & \multicolumn{2}{c}{\textbf{Qasper}} & \multicolumn{3}{c}{\textbf{Overall}} \\
\cmidrule(lr){3-5} \cmidrule(lr){6-7} \cmidrule(lr){8-10} \cmidrule(lr){11-12} \cmidrule(lr){13-15}
 & \textbf{Method} & Acc & Rec & NLI & Acc & NLI & Acc & Rec & NLI & Rec & NLI & Acc & Rec & NLI \\
\midrule

& GQA & 71.6 & 49.0 & 23.8 & 80.7 & 27.2 & 61.8 & 66.9 & 50.6 & 65.8 & 37.5 & 71.4 & 60.6 & 34.8 \\ 
& CoT & 71.1 & 50.9 & 26.2 & 85.1 & 46.5 & 67.4 & \textbf{68.1} & 49.0 & 64.9 & 46.2 & \textbf{74.5} & 61.3 & 42.0 \\
& PS & 67.4 & 46.6 & 31.2 & 81.1 & 42.2 & 65.2 & 65.5 & 41.6 & 62.4 & 40.5 & 71.2 & 58.2 & 38.9 \\
& ReAct & 58.0 & 30.8 & 12.3 & 79.2 & 11.2 & 44.2 & 40.8 & 10.1 & 49.8 & 24.1 & 60.5 & 40.5 & 14.4 \\
& PEARL & 70.3 & 50.9 & 38.1 & \textbf{85.3} & \textbf{53.2} & 56.8 & 64.5 & \textbf{72.1} & 66.5 & 51.1 & 70.8 & 60.6 & 53.6 \\ 
\cdashline{2-15}
\rowcolor{softblue} 
\cellcolor{white} \multirow{-6}{*}{\rotatebox{90}{\shortstack[c]{\textbf{GPT-4o} \\ \textbf{mini}}}}
& \textbf{\alg} & \textbf{73.4} & \textbf{54.0} & \textbf{41.0} & 81.3 & 50.3 & \textbf{67.6} & 66.6 & 69.9 & \textbf{67.1} & \textbf{61.8} & 74.1 & \textbf{62.6} & \textbf{55.8} \\
\midrule

& GQA & 68.5 & 47.7 & 23.2 & 77.7 & 21.4 & 59.2 & 57.7 & 37.1 & 60.5 & 31.5 & 68.5 & 55.3 & 28.3 \\ 
 & CoT & 68.1 & 46.0 & 16.5 & \textbf{82.2} & 39.6 & 55.8 & 55.0 & 29.5 & 61.3 & 36.6 & 68.7 & 54.1 & 30.6 \\
 & PS & 66.7 & 44.7 & 16.9 & 80.8 & 25.7 & 48.3 & 42.0 & 33.1 & 58.6 & 36.2 & 65.3 & 48.4 & 28.0 \\
 & ReAct & 42.4 & 19.7 & 12.3 & 71.0 & 6.3 & 43.4 & 40.8 & 27.9 & 25.1 & 18.9 & 52.3 & 28.5 & 16.4 \\
 & PEARL & 68.5 & 51.4 & 48.5 & 80.1 & \textbf{75.9} & 55.8 & 56.9 & 76.2 & 66.0 & 74.8 & 68.1 & 58.1 & 68.9 \\ 
\cdashline{2-15}
\rowcolor{softblue} 
\cellcolor{white} \multirow{-6}{*}{\rotatebox{90}{\shortstack[c]{\textbf{Llama} \\ \textbf{3.1-8B}}}}
& \textbf{\alg} & \textbf{70.8} & \textbf{53.9} & \textbf{50.5} & 79.1 & \textbf{75.9} & \textbf{67.9} & \textbf{62.3} & \textbf{77.9} & \textbf{67.3} & \textbf{75.6} & \textbf{72.6} & \textbf{61.2} & \textbf{70.0} \\
\midrule

& GQA & 70.8 & 44.6 & 21.0 & 78.9 & 17.8 & 64.1 & 63.2 & 40.2 & 62.5 & 34.9 & 71.3 & 56.8 & 28.5 \\ 
 & CoT & 68.3 & 45.3 & 22.1 & 83.1 & 36.7 & 67.0 & 64.9 & 43.2 & 63.4 & 36.6 & 72.8 & 57.9 & 34.7 \\
 & PS & 69.5 & 42.6 & 26.5 & 81.5 & 33.9 & 61.1 & 60.5 & 30.9 & 62.8 & 39.0 & 70.7 & 55.3 & 32.6 \\
 & ReAct & 59.9 & 32.0 & 13.2 & 80.9 & 11.5 & 51.0 & 49.1 & 15.6 & 41.5 & 25.1 & 63.9 & 40.9 & 16.4 \\
 & PEARL & 72.7 & 48.4 & 37.7 & 81.4 & \textbf{56.1} & 60.1 & 61.2 & 66.6 & 63.6 & 46.1 & 71.4 & 57.7 & 51.6 \\ 
\cdashline{2-15}
\rowcolor{softblue} 
\cellcolor{white} \multirow{-6}{*}{\rotatebox{90}{\shortstack[c]{\textbf{Qwen} \\ \textbf{2.5-14B}}}}
& \textbf{\alg} & \textbf{75.1} & \textbf{49.8} & \textbf{39.6} & \textbf{83.9} & 54.8 & \textbf{72.3} & \textbf{66.3} & \textbf{68.9} & \textbf{67.1} & \textbf{56.4} & \textbf{77.1} & \textbf{61.1} & \textbf{54.9} \\
\bottomrule
\end{tabular}
}

\caption{Performance comparison on various datasets. \textbf{Bold} indicates the best performance.} 
\label{tab:main_results}

\end{table*}

\paragraph{Evaluation Metrics.} We primarily calculate token-level recall~\cite{lin-2004-rouge} and Natural Language Inference (NLI) scores~\cite{honovich-etal-2022-true-evaluating, chen-eger-2023-menli} to evaluate the quality of generated answers from multiple perspectives. 
NLI-based evaluation~\cite{chen-eger-2023-menli} assesses semantic equivalence through logical entailment, effectively capturing nuances in noisy reasoning sequences where surface-level string matching often fails~\cite{balamurali2025revisitingnlicosteffectivehumanaligned}.
Our NLI-based evaluation uses the entailment probability from DeBERTa-V3-Large~\citep{he2023debertav3}. To identify the presence of correct information within long answers, we follow the granularity alignment strategy~\cite{laban2022summac} by applying a sliding window and selecting the maximum entailment probability as the final score.
For ease of interpretation and consistency with accuracy, NLI scores and Recall are scaled to a 0--100 range.

For ConditionalQA, where we focus on the binary (Yes/No) classification, we exclude recall evaluation because the highly restricted answer format makes token-level metrics inappropriate.
For NLI-based evaluation, we format each question and its ground truth into a declarative statement (e.g., 'The answer to [X] is [Y]') to verify whether generated responses logically entail the correct answer.

For multiple-choice datasets, we measure accuracy using GPT-4o as a judge~\citep{zheng2023judging} to map the generated responses to the available options, as it shows a high correlation with human judgment and identifies semantic equivalence more accurately than simple string matching.
\autoref{tab:LLM-as-a-Judge} and Appendix~\ref{sec:appendix_exp_details} provide implementation details. 
For Qasper, accuracy evaluation is excluded as the dataset primarily consists of free-form responses rather than multiple-choice questions.

\paragraph{Baselines.} We compare \alg with the following baselines: Generative Question Answering (GQA), zero-shot Chain-of-Thought (CoT)~\cite{wei2022chain}, Plan-and-Solve~\cite{wang-etal-2023-plan}, ReAct~\cite{yao2023react}, and PEARL~\cite{sun-etal-2024-pearl}.
For CoT, Plan-and-Solve, and ReAct, we evaluate the full reasoning process, relying on NLI-based metrics to mitigate potential recall bias. To ensure a consistent comparison, our method \alg and ReAct adopts the same action space as PEARL. For both PEARL and \alg, we set the plan correction budget $B=7$ to balance performance and inference efficiency. We set the maximum number of negative constraints to $k=3$. Specific prompts are provided in Table \ref{tab:GenerativeQA}, \ref{tab:CoT}, \ref{tab:Plan-and-Solve}, \ref{tab:react_action}, and \ref{tab:react_execution}.
For both PEARL and the Executor of \alg, we adopt the original prompt configurations as specified in \citet{sun-etal-2024-pearl}.

\subsection{Results}

\autoref{tab:main_results} shows that \alg consistently achieves competitive performance across models. It yields notable gain over PEARL on Llama-3.1-8B-Instruct (+4.5\% acc, +3.1\% recall) and achieves the best accuracy on Qwen-2.5-14B-Instruct (77.1\%).
The improvements in the NLI metric, which evaluates the validity of reasoning, are even more significant than those in standard accuracy. Compared to CoT, \alg{} increases overall NLI scores by 13.8 points for GPT-4o-mini and 20.2 points for Qwen-2.5-14B-Instruct. The results on Llama are particularly remarkable; our method raises the NLI score from 30.6 to 70.0, representing a more than two-fold increase. This substantial performance gap demonstrates that \alg{} identifies logical causalities via constraints rather than merely matching keywords or generating superficially plausible yet logically flawed sentences.

The corrective impact of this framework is most pronounced in open-source models with relatively limited reasoning capabilities, such as Llama and Qwen. Notably, Qwen achieved the highest accuracy of 77.1\% across all experimental groups, showing that our approach effectively unlocks the model's latent reasoning potential. These results suggest that the proposed method is a robust solution for maximizing the latent reasoning capacity within small to medium-sized models.

On the ConditionalQA dataset, \alg{} occasionally falls short of the highest scores in specific metrics. This gap likely stems from the multi-faceted analysis of our planning process, which increases information richness to avoid bias. The NLI model often perceives this multi-faceted content as noise or information dilution, resulting in underestimated quantitative scores. Appendix~\ref{sec:appendix_faith} provides a detailed analysis and quantitative proof of this phenomenon.
\definecolor{softblue}{RGB}{240, 245, 255}
\begin{table}[t]
\centering
\resizebox{\columnwidth}{!}{
\begin{tabular}{l ccc ccc}
\toprule
& & & & \multicolumn{3}{c}{\textbf{LongReason}} \\
\cmidrule(lr){5-7}
\textbf{Model} & $\mathcal{M}_{pred}$ & $\mathcal{M}_{plan}$ & $\mathcal{M}_{corr}$ & Acc & Rec & NLI \\
\midrule

\rowcolor{softblue} \cellcolor{white} & \cmark & \cmark & \cmark & \textbf{60.9} & \textbf{61.9} & \textbf{65.6} \\
\cdashline{2-7}
\cellcolor{white} & \cmark & \cmark & \xmark & 47.5 & 48.8 & 51.4 \\
\cellcolor{white} & \cmark & Van.   & \xmark & 53.1 & 54.8 & 60.9 \\
\cellcolor{white} \multirow{-4}{*}{\rotatebox[origin=c]{90}{\makecell{\textbf{GPT-4o} \\ \textbf{mini}}}} & \xmark  & Van.   & \xmark & 37.7 & 42.7 & 48.9 \\
\midrule

\rowcolor{softblue} \cellcolor{white} & \cmark & \cmark & \cmark & \textbf{59.8} & \textbf{57.0} & \textbf{60.4} \\
\cdashline{2-7}
\cellcolor{white} & \cmark & \cmark & \xmark & 50.6 & 47.2 & 50.5 \\
\cellcolor{white} & \cmark & Van.   & \xmark & 48.5 & 43.7 & 53.0 \\
\cellcolor{white} \multirow{-4}{*}{\rotatebox[origin=c]{90}{\makecell{\textbf{Qwen} \\ \textbf{2.5-14B}}}} & \xmark & Van. & \xmark & 40.9 & 42.2 & 47.5 \\
\bottomrule
\end{tabular}
}
\caption{Ablation study on the LongReason dataset using GPT-4o-mini and Qwen-2.5-14B-Instruct. We evaluate the contribution of each component. Van. denotes Vanilla, representing a standard planner without the strategy reasoning process.
}
\label{tab:ablation}

\end{table}
\begin{figure}[t]
\centering
\includegraphics[width=0.5\textwidth]{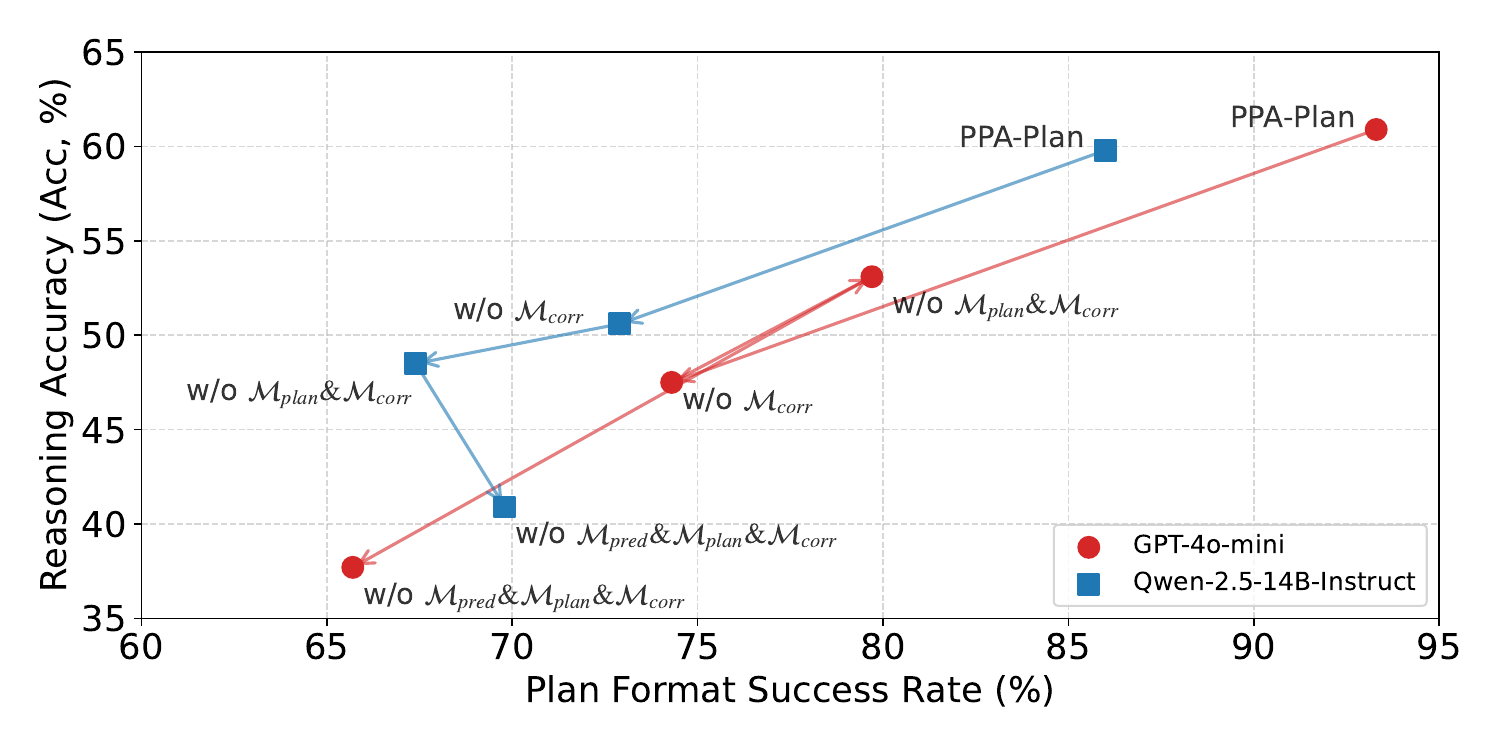}
\caption{
Impact of \alg{} components on plan executability and reasoning accuracy.
}
\label{fig:scatter}
\end{figure}

\subsection{Ablation Study}
This section analyzes how each component (the Pitfall Predictor $\mathcal{M}_{pred}$, the Constraint-Aware Planner $\mathcal{M}_{plan}$, and the Context-Aware Corrector $\mathcal{M}_{corr}$) contributes to the overall performance of \alg{}. \autoref{tab:ablation} shows that for both GPT-4o-mini and Qwen-2.5-14B-Instruct, the full \alg{} configuration yields the best results across all metrics. We adopt the original prompt configurations for the Vanilla Planner, as specified in \citet{sun-etal-2024-pearl}.

Removing $\mathcal{M}_{pred}$ in the vanilla planner setup also results in performance degradation. For GPT-4o-mini, accuracy drops by 15.4\% (from 53.1\% to 37.7\%) and the NLI score by 12 (from 60.9 to 48.9). A similar trend is observed in Qwen-2.5-14B-Instruct, where accuracy and NLI score decrease by 7.6\% (from 48.5\% to 40.9\%) and by 5.5 (from 53.0 to 47.5). These results suggest that a planner struggles to identify logical pitfalls on its own. Thus, preemptive pitfall identification provides a critical guide for the planning process.

\begin{figure}[t]
\centering
\includegraphics[width=0.5\textwidth]{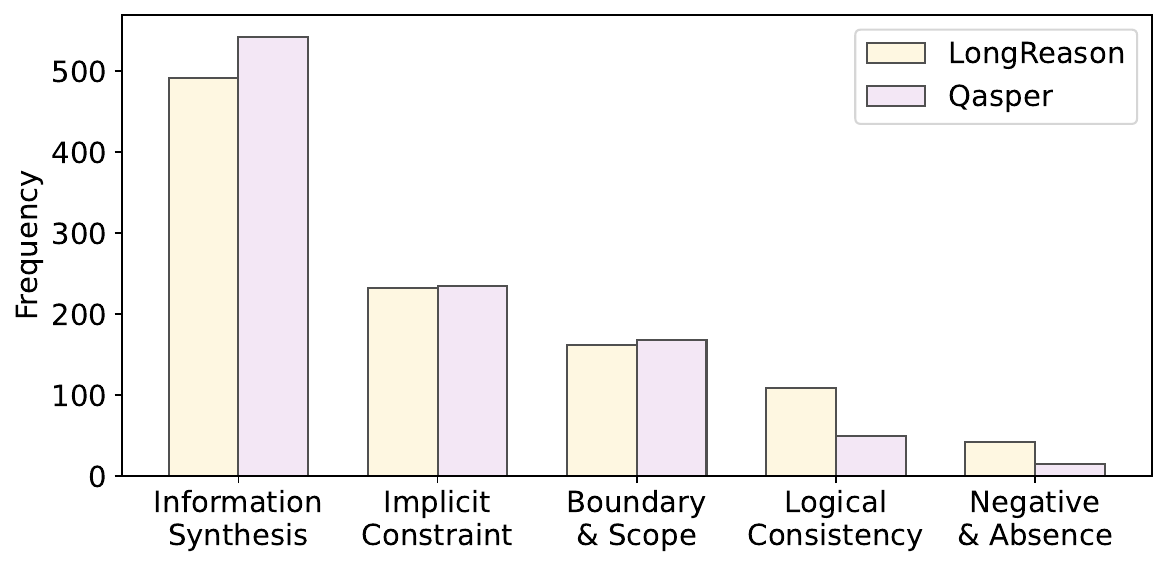}
\caption{
Distribution of negative constraint types generated by the Pitfall Predictor on LongReason and Qasper.
}
\label{fig:neg_const}
\end{figure}

\begin{figure*}[t]
\centering
\begin{subfigure}[t]{0.48\textwidth}
\includegraphics[width=\textwidth]{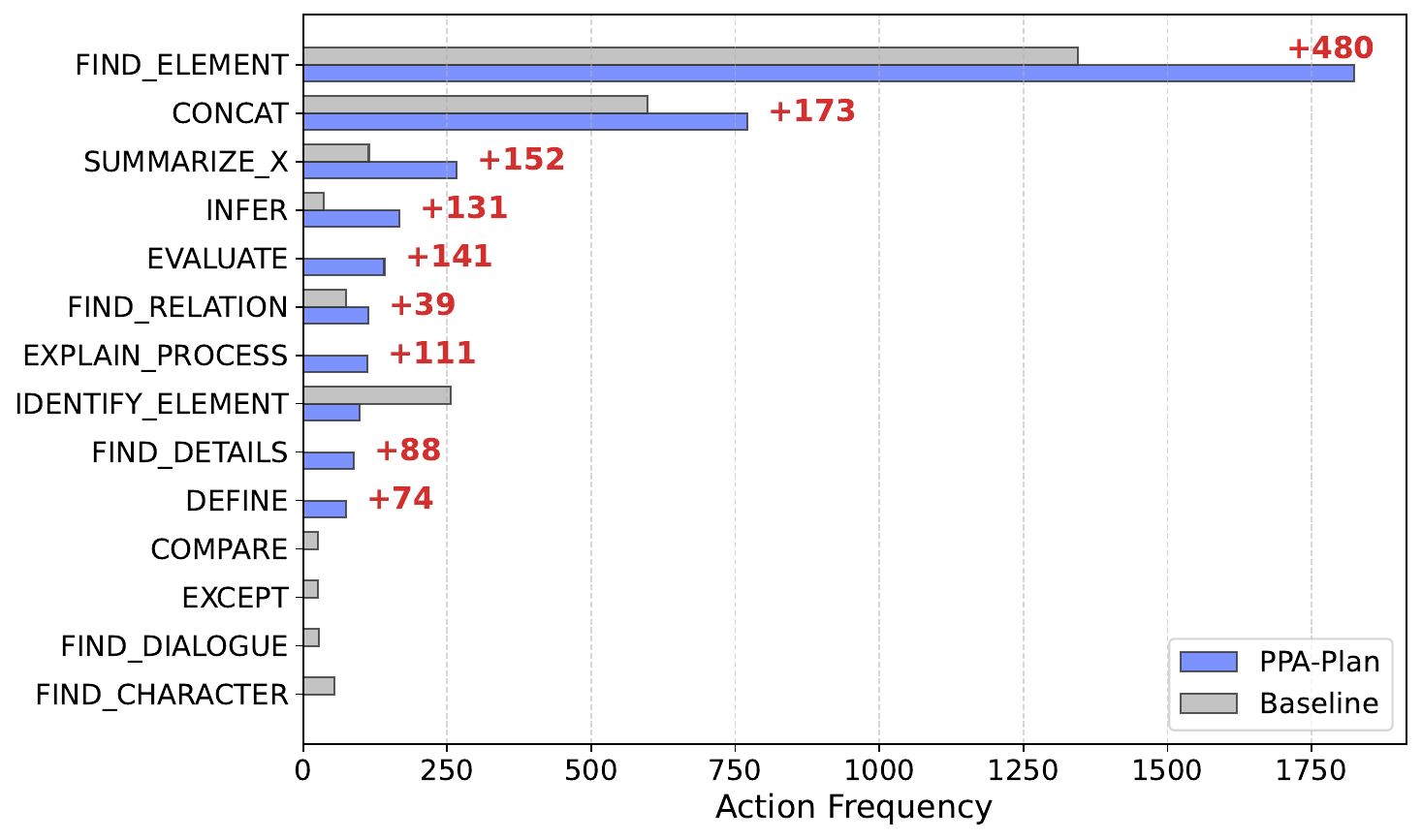}
\caption{
LongReason
}
\label{fig:act_anl_lr}
\end{subfigure}
\hfil
\begin{subfigure}[t]{0.48\textwidth}
\includegraphics[width=\textwidth]{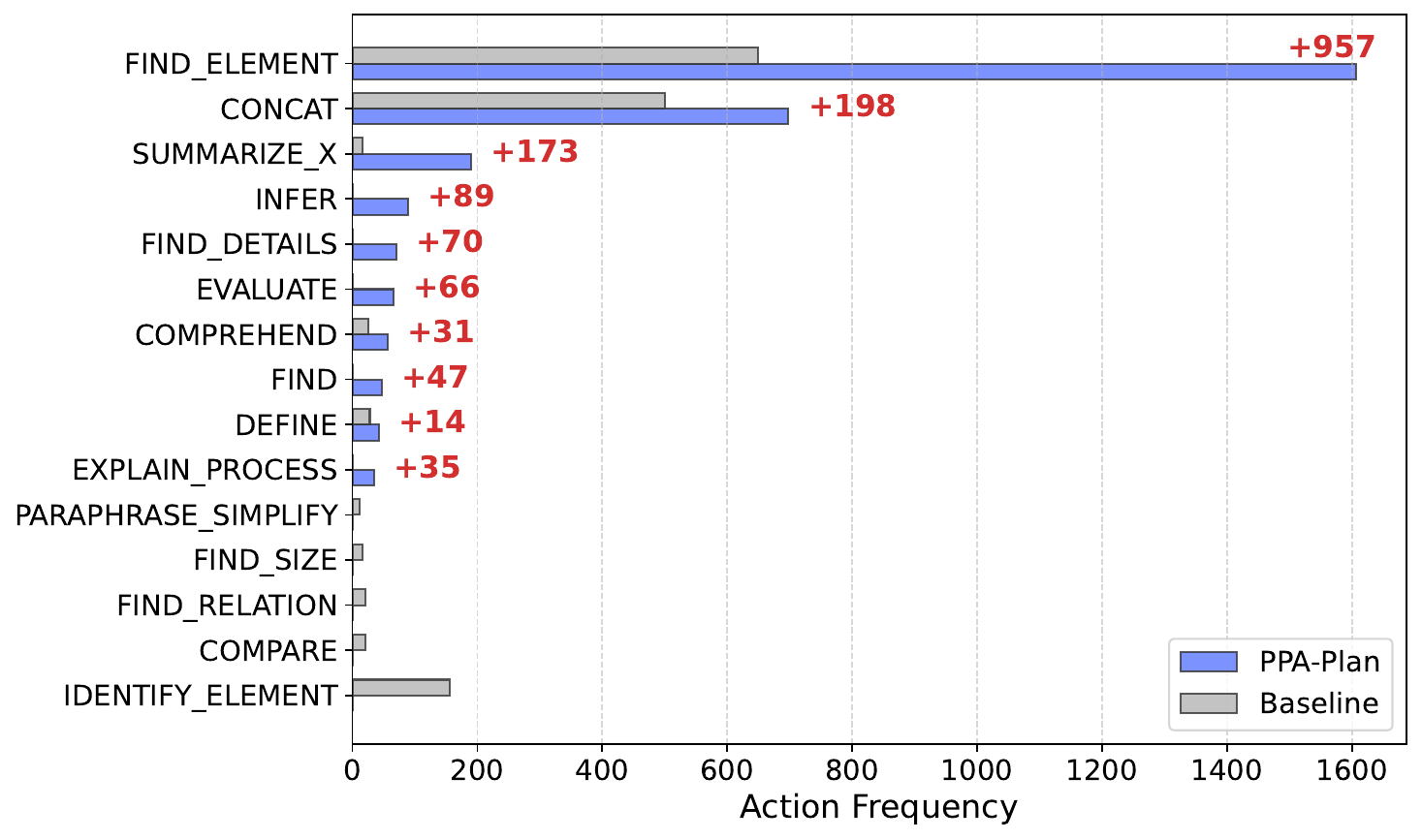}
\caption{
Qasper
}
\label{fig:act_anl_qas}
\end{subfigure}
\caption{
Strategic shift in action distributions induced by negative constraints and strategy reasoning. 
The baseline refers to a vanilla plan-and-execute setup that uses the Vanilla Planner without negative constraints and strategy reasoning, in contrast to \alg{}, which incorporates both.
}
\label{fig:act_anl}
\end{figure*}

Removing $\mathcal{M}_{corr}$ also leads to a substantial performance drop, demonstrating its role as a safety net for plan executability. For GPT-4o-mini, accuracy decreases by 13.4\% and the NLI score by 14.2; similarly, Qwen-2.5-14B-Instruct exhibits declines of 9.2\% and 9.9.
Interestingly, GPT-4o-mini performs worse when removing only $\mathcal{M}_{corr}$ (47.5\%) than when removing the proposed $\mathcal{M}_{plan}$ and $\mathcal{M}_{corr}$ (53.1\%). As shown in \autoref{fig:scatter}, this occurs because sophisticated plans from $\mathcal{M}_{plan}$ often lead to format violations~\cite{saparov2023language, zhao2025trade}. Without $\mathcal{M}_{corr}$, the success rate of valid formats falls to 74.3\%, lower than the baseline's 79.7\%, whereas the full framework reaches 93.3\%. This underscores that $\mathcal{M}_{\mathrm{corr}}$ is essential for maintaining the validity of complex plans, creating an organic synergy between the proposed modules.

Conversely, Qwen shows an increased success rate when removing only $\mathcal{M}_{corr}$ (72.9\%) compared to the baseline, where the proposed $\mathcal{M}_{plan}$ and $\mathcal{M}_{corr}$ are absent (67.4\%). This trend indicates that the plan generated by Qwen’s $\mathcal{M}_{plan}$ has relatively low complexity, which a baseline corrector can handle to some extent. These findings confirm that $\mathcal{M}_{corr}$ becomes increasingly vital for high-performance models; because these models generate more sophisticated plans, they require a specialized correction module to ensure executability without violating the required format.

\subsection{Analysis}
\label{sec:analysis}

\paragraph{Distribution of Negative Constraints.}
We analyzed the characteristics of negative constraints by classifying 200 sampled instances from LongReason and Qasper using GPT-4o across five categories. As shown in Figure~\ref{fig:neg_const}, information synthesis is the most frequent type, followed by implicit constraints and boundary \& scope. This distribution confirms that our framework identifies pitfalls requiring global information integration and logical depth beyond surface-level text. Full definitions and statistics are provided in Appendix~\ref{sec:appendix_neg_const}.

\paragraph{Impact on Planning Behavior.} This section examines how negative constraints transform the planner’s reasoning. We compare \alg{} with a baseline that uses the Vanilla Planner without negative constraints and strategy reasoning using Qwen-2.5-14B-Instruct on LongReason and Qasper, focusing on the top 10 actions. Given the consistent patterns across both datasets, we use LongReason as a representative case.

Results show that applying negative constraints induces substantial changes in the model’s reasoning process. Most notably, this approach improves plan complexity and sophistication. 
The average plan step count increases by 1.26 from 4.47 to 5.73 in LongReason and by 1.45 from 2.98 to 4.43 in Qasper compared to the baseline. 
This increase indicates that the model establishes more detailed and sophisticated plans to preemptively block potential errors by using negative constraints and strategy reasoning.

Second, action patterns shift from simple extraction toward logical reasoning. \autoref{fig:act_anl} shows that high-level reasoning actions, such as \texttt{INFER} and \texttt{SUMMARIZE\_X}, increase significantly. This trend suggests that constraints force the model to logically connect and summarize gathered information instead of simply listing it.

Third, the model’s self-verification mechanism becomes active. \texttt{EVALUATE} and \texttt{EXPLAIN\_PROCESS} appear frequently in plans using constraints, even though they were absent from the top baseline actions. This shift shows that the model does more than verify whether its plan satisfies the given constraints;
it also facilitates a more objective assessment of the context, thereby mitigating potential reasoning biases.

Fourth, information-collection behaviors have advanced significantly. The frequency of entity search actions relying on surface-level information, such as \texttt{FIND\_CHARACTER}, \texttt{FIND\_DIALOGUE}, \texttt{COMPARE}, and \texttt{IDENTIFY\_ELEMENT}, decreases. In contrast, deep evidence collection actions such as \texttt{FIND\_ELEMENT} and \texttt{FIND\_DETAILS} increase, helping the model collect specific evidence, identify logical contradictions, and explore logical interdependencies among various elements. This contrast indicates that the model avoids surface-level keyword searches and instead collects multifaceted logical evidence to improve the precision of its reasoning.

\paragraph{Plan Execution Failure.} 
To understand failures in plan generation, we summarize plan execution failure rates (\%) across all datasets for PEARL and \alg.
\begin{table}[t]
\centering
\resizebox{\columnwidth}{!}{
\begin{tabular}{lc cccc}
\toprule
 & \textbf{Method} & \textbf{QuALITY} & \textbf{Cond.QA} & \textbf{LongReason} & \textbf{Qasper} \\
\midrule

\multirow{2}{*}{\makecell[c]{\textbf{GPT-4o}\\ \textbf{mini}}}
& PEARL & 27.4 & 9.4 & 34.3 & 45.9 \\
& \textbf{\alg} & \textbf{3.1} & \textbf{0.4} & \textbf{6.7} & \textbf{1.0} \\
\midrule

\multirow{2}{*}{\makecell[c]{\textbf{Llama}\\ \textbf{3.1-8B}}}
& PEARL & 39.4 & 60.0 & 62.0 & 65.6 \\
& \textbf{\alg} & \textbf{19.7} & \textbf{22.8} & \textbf{35.1} & \textbf{23.2} \\
\midrule

\multirow{2}{*}{\makecell[c]{\textbf{Qwen}\\ \textbf{2.5-14B}}}
& PEARL & 17.4 & 20.7 & 30.7 & 38.2 \\
& \textbf{\alg} & \textbf{13.3} & \textbf{5.0} & \textbf{14.0} & \textbf{11.2} \\
\bottomrule
\end{tabular}
}
\caption{Plan execution failure rates (\%) of PEARL and \alg across different base models and datasets.
}

\label{tab:plan_failure_analysis}

\end{table}
As shown in \autoref{tab:plan_failure_analysis}, \alg consistently reduces the failure rate compared to PEARL. Specifically, when using GPT-4o-mini on the Qasper dataset, \alg lowers the failure rate from 45.9\% to 1.0\%. This significant reduction shows that \alg achieves superior plan executability while maintaining high quality.

\begin{table}[t]
\centering
\small
\begin{tabular}{l c}
\toprule
\textbf{Metric} & \textbf{Value (\%)} \\
\midrule
\multicolumn{2}{l}{\textit{Constraint Quality}} \\
Ungrounded Constraint Rate & 31.89 \\
Harmful Constraint Rate & 21.93 \\
\midrule
\multicolumn{2}{l}{\textit{Accuracy by Constraint Validity}} \\
Invalid Constraints & 64.58 \\
Valid Constraints & 71.22 \\
Overall Accuracy & 69.10 \\
\bottomrule
\end{tabular}
\vspace{-2mm}
\caption{Quality of generated negative constraints and task accuracy under valid and invalid constraint conditions.
}
\vspace{-2mm}
\label{tab:const_quality}
\end{table}
\paragraph{Quality and Robustness of Negative Constraints.}
To address the quality of negative constraints and the potential risks of $\mathcal{M}_{pred}$, we analyzed 300 negative constraints generated by $\mathcal{M}_{pred}$ (Qwen-2.5-14B-Instruct) using 100 random samples from the LongReason dataset. We used Claude-4.5-Sonnet as a judge to label a negative constraint as \textit{ungrounded} when it warns against a risk not evidenced in the context, and as \textit{harmful} when it misdirects the correct reasoning process.
As shown in \autoref{tab:const_quality}, 31.89\% of the generated constraints are ungrounded and 21.93\% are harmful. 
This indicates that, even with some noise, the constraints function as useful structural cues rather than strict supervision signals.
To further evaluate the robustness of \alg against imperfect constraint generation, we analyze how invalid constraints affect task performance. We define a negative constraint as \textit{invalid} if it is either \textit{ungrounded} or \textit{harmful}, and measure task accuracy separately under valid and invalid constraints. Accuracy remains relatively robust at 64.58\%, representing a drop of only 6.64\% compared to performance with valid constraints (71.22\%). This result suggests that \alg remains robust despite these invalid constraints.

We also conduct failure category analysis, where we use an LLM judge to identify the specific causes of invalid constraints, allowing multiple labels per instance to capture overlapping issues. As shown in \autoref{tab:const_error_cat}, \textit{Structural Hallucination} (63.3\%) and \textit{Over-thinking} (48.0\%) are the primary issues. These errors occur because $\mathcal{M}_{pred}$ detects potential pitfalls based only on the question to maximize efficiency before reviewing the context. This design encourages the model to generate conservative negative constraints for extreme scenarios.

\begin{table}[t]
\centering
\small
\begin{tabular}{l c}
\toprule
\textbf{Error Category} & \textbf{Rate (\%)} \\
\midrule

Structural Hallucination & 63.3 \\
Over-thinking & 48.0 \\
Logic/Format Misunderstanding & 24.5 \\
Others & 12.2 \\
\bottomrule
\end{tabular}
\vspace{-2mm}
\caption{Error categories of invalid negative constraints. Multiple labels are allowed per instance.
}
\label{tab:const_error_cat}

\end{table}

\section{Conclusion}

To address the tendency of LLM-based planners to anchor on initial flawed outputs during reactive refinement, we propose \alg{}, a planning strategy that proactively prevents logical pitfalls in planning by introducing negative constraints prior to plan generation. 
Experimental results demonstrate that \alg{} consistently achieves higher accuracy and stronger long-context reasoning capabilities than existing baselines. 
Our analysis further reveals that, by incorporating negative constraints with multiple perspectives, \alg{} encourages more deliberate planning behaviors, shifting from surface-level keyword search toward analytical actions such as evaluation and reasoning.
In conclusion, this study establishes a systematic framework that allows LLMs to recognize pitfalls and think strategically before execution. 

\section*{Limitations}
While \alg{} achieves impressive results, we identify the following limitations. First, because the Context-Aware Corrector assumes the model can generate a baseline level of coherent output, small-scale models struggle to generate a valid plan format, despite the correction process. Second, the framework operates on the premise that negative constraints accurately capture actual pitfalls. Identifying a false pitfall introduces noise that disrupts the planning process, which prevents the model from establishing an accurate plan. Third, our system inherits the efficiency issues found in PEARL; repeatedly processing long contexts alongside multiple plan steps results in slow inference speeds.
To strengthen system robustness, we plan to explore fine-tuning techniques for smaller models and add a module to verify the logical validity of generated negative constraints, along with practical utility by applying context caching for inference efficiency.

\section*{Acknowledgments}
This work was supported by the National Research Foundation of Korea (NRF) grant funded by the Korea government (MSIT) (RS-2025-24535182 and RS-2026-25498006).

\bibliography{custom}
\newpage
\appendix

\clearpage
\section{Appendix}

\subsection{Experimental Setup Details}
\label{sec:appendix_exp_details}
All local experiments were conducted on a single NVIDIA A6000 48GB GPU.
For all models, including GPT-4o (LLM-as-a-Judge), we employed greedy decoding (temperature = 0, do\_sample = False) to ensure reproducibility. 
For all generation tasks, we set the maximum number of output tokens to 512, whereas the pitfall predictor was limited to 256 tokens.
In the NLI-based evaluation, we set the window size to 512 and stride to 256.
Since all method is training-free and requires no parameter updates, we utilized the entire set of available samples for evaluation, comprising the original training, validation, and test splits, to ensure statistical robustness.
Adopting the setup from PEARL~\cite{sun-etal-2024-pearl}, we utilized the human annotation scores to distinguish task difficulty. An average score $\ge 3$ were classified into the long split, while the remainder were classified into the short split.
The detailed composition of our evaluation dataset is in \autoref{tab:dataset_details}.
\begin{table}[h]
\centering
\resizebox{\columnwidth}{!}{
\begin{tabular}{l c c c}
\toprule
\textbf{Dataset} & Config. & Split & Samples \\
\midrule

& long & dev & 330 \\
& long & train & 368 \\
& short & dev & 302 \\
\cdashline{2-4}
\multirow{-4}{*}{\rotatebox[origin=c]{0}{QuALITY}} &  &  & 1000 \\
\midrule

& binary & train & 860 \\
& binary & dev & 140 \\
\cdashline{2-4}
\multirow{-3}{*}{\rotatebox[origin=c]{0}{Cond.QA}} & & & 1000 \\
\midrule

& 16k & - & 500 \\
& 32k & - & 500 \\
\cdashline{2-4}
\multirow{-3}{*}{\rotatebox[origin=c]{0}{LongReason}} &  &  & 1000 \\
\midrule

& free-form & dev & 300 \\
& free-form & train & 600 \\
\cdashline{2-4}
\multirow{-3}{*}{\rotatebox[origin=c]{0}{Qasper}} &  &  & 900 \\
\bottomrule
\end{tabular}
}
\caption{Overall composition and statistics of the evaluation datasets used for the main results. Cond.QA denotes ConditionalQA dataset, Config. denotes Configuration.
}
\label{tab:dataset_details}

\end{table}

\begin{figure}[h]
    \centering
    \includegraphics[width=0.5\textwidth]{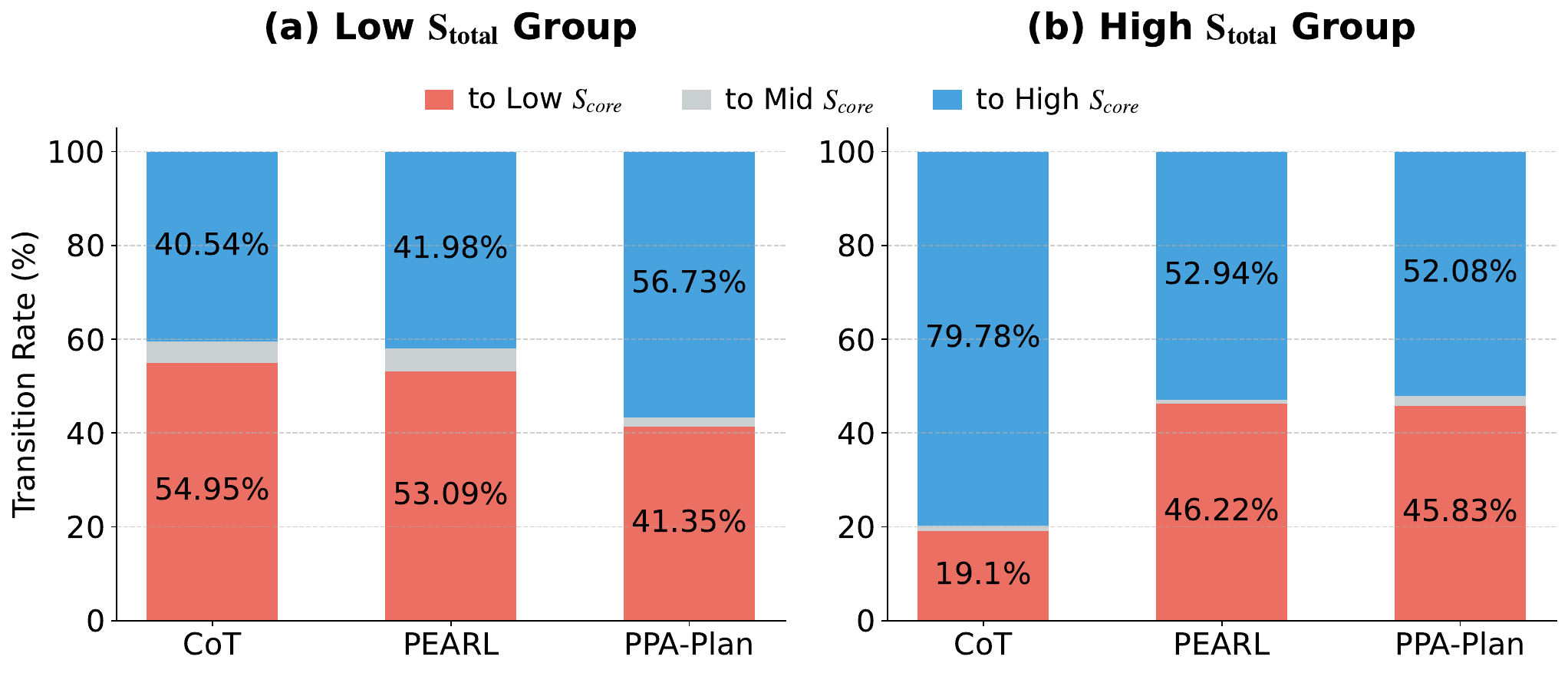}
    \caption{NLI transition analysis of $S_\mathrm{total}$ to $S_\mathrm{core}$. (a) represents the recovery rate in the low-score group, while (b) shows the logical density and evidence retention in the high-score group. Note that \alg{} successfully bypasses potential pitfalls through multi-faceted reasoning.}
    \label{fig:nli_transition}
\end{figure}
\subsection{Intrinsic Reasoning Faithfulness}
\label{sec:appendix_faith}
To verify the logical robustness and correctness of the answers generated under our constraint sets, we conducted a separate core conclusion extraction analysis. As shown in \autoref{fig:nli_transition}, using 200 randomly sampled instances from the ConditionalQA dataset, we compared the original answer scores ($S_{\mathrm{total}}$) with the scores of core conclusions ($S_\mathrm{core}$) extracted via GPT-4o-mini. We categorized these instances into low-score ($S_\mathrm{total} < 0.3$) and high-score ($S_\mathrm{total} > 0.7$) groups to measure the transition rates between these score states. 
Experimental results show that \alg{} achieves a 56.73\% transition rate from the initial low $S_\mathrm{total}$ group to the high $S_\mathrm{core}$ group. This rate is more than 15\% higher than those of CoT (40.54\%) and PEARL (41.98\%). This result proves that the abundance of information in \alg{}, generated to avoid bias or perform multi-perspective analysis, acts as noise during NLI evaluation rather than indicating actual logical failures. While \alg{} maintains logical robustness and correctness through deep reasoning, NLI metrics underestimate its performance more severely than other methods. In contrast, CoT shows the lowest transition rate, suggesting its low scores likely stem from actual logical flaws rather than information richness.

Analysis of the high-score ($S_\mathrm{total}$) group reveals that extracting core conclusions caused transitions to low or mid $S_\mathrm{core}$ at rates of 47.06\% for PEARL and 47.91\% for \alg{}. These figures represent a score drop approximately 27\% larger than the 20.22\% observed for CoT. This significant decrease indicates that the detailed explanations generated by \alg{} serve as essential logical grounding that supports the final answer. Notably, CoT exhibited the highest state-retention rates, 54.95\% for low-to-low and 79.78\% for high-to-high, confirming that NLI metrics underestimate CoT the least among all compared approaches.

\subsection{Distribution of Negative Constraints.}
\label{sec:appendix_neg_const}
This section analyzes the error types identified by the Pitfall Predictor and the characteristics of the resulting negative constraints. We randomly sampled 200 instances of negative constraints generated using Qwen-2.5-14B-Instruct from the LongReason and Qasper datasets. GPT-4o then served as an LLM judge to perform multi-label classification across five predefined categories (Figure~\ref{fig:neg_const}).

Results show that information synthesis is the most frequent category in both datasets, with 491 labels in LongReason and 542 in Qasper. The total label count exceeds the number of samples because a single negative constraint can belong to multiple categories. This high frequency indicates that our constraints force the model to integrate scattered information throughout long contexts during reasoning. By requiring this synthesis, the framework prevents the model from drawing conclusions based solely on localized information within specific paragraphs.

Implicit constraints also account for a significant portion of the labels (232 and 235 occurrences, respectively). These constraints address essential preconditions for logical completeness not explicitly stated in the text. By identifying these hidden requirements, the negative constraints prevent models from merely reacting to surface-level text and instead force them to recognize underlying logical constraints for deeper reasoning.

Finally, the boundary \& scope category (162 and 168 occurrences) optimizes the vast search space within the specific limits of the problem. This approach prevents premature generalizations or unfounded logical expansions, where models create unfounded causal links to jump to conclusions, allowing the model to maintain reasoning precision within the prescribed domain.

\subsection{Token Efficiency Analysis.}
To assess whether the performance gains of \alg are simply due to a larger token budget, we report token statistics for LongReason (16k) with GPT-4o-mini.

\begin{itemize}
    \item PEARL: average total tokens = 8206.84, average calls = 1.12
    \item PPA-Plan: average total tokens = 7275.11, average calls = 1.44
\end{itemize}

Despite involving an additional prediction step instead of an additional long-context reasoning step, \alg uses fewer total tokens on average than PEARL (7275 vs. 8206 tokens). The increase in the number of calls does not translate into a larger overall token budget. Instead, \alg replaces full-length regeneration with short negative constraint guidance, resulting in more efficient corrections. Therefore, the performance improvements shown in \autoref{tab:main_results} cannot be attributed to increased token usage. Instead, they stem from the structural guidance introduced by negative constraints within the pipeline design.

\subsection{Sensitivity of Evaluation Protocol.}
To assess whether our results are sensitive to specific evaluation configurations, we diversified the evaluation protocol as follows: (i) using a different LLM-judge (i.e., Claude 4.5 Sonnet instead of GPT-4o), (ii) using an alternative prompt configuration for the GPT-4o judge (i.e., a structured framing focused on factual consistency with the provided evidence rather than general accuracy), (iii) using a BART-large-mnli~\cite{lewis-etal-2020-bart} evaluator to ensure logical consistency beyond simple mapping. We compare one competitive baseline PEARL, and our \alg on 200 sampled instances from LongReason using Qwen-2.5-14B-Instruct as the base model, as shown in \autoref{tab:eval_sensitivity}.
\begin{table}[h]
\centering
\resizebox{\columnwidth}{!}{
\begin{tabular}{l c c c}
\toprule
\textbf{Method} 
& \makecell{\textbf{Claude 4.5} \\ \textbf{Sonnet} (Acc.)}
& \makecell{\textbf{GPT-4o Alternative} \\ \textbf{Prompt} (Acc.)}
& \makecell{\textbf{BART-large-mnli} \\ (NLI Score)} \\
\midrule

PEARL & 54.5 & \textbf{69.0} & 44.4 \\
\textbf{\alg} & \textbf{60.5} & \textbf{69.0} & \textbf{51.0} \\
\bottomrule
\end{tabular}
}
\caption{Performance of PEARL and \alg under alternative evaluation protocols, including different judges, prompts, and evaluators, on 200 sampled LongReason instances.
}
\label{tab:eval_sensitivity}

\end{table}

Despite using different judges (i), prompts (ii), and evaluators (iii), \alg consistently performs at or above the level of PEARL. Interestingly, we observe that Claude 4.5 Sonnet refuses to select an option if the open-ended answer lacks a strict match, whereas GPT-4o tends to choose one of the available options regardless. This strictness in Claude’s criteria results in overall lower accuracy scores compared to GPT-based evaluations, yet \alg maintains its relative advantage or parity in both settings.

\clearpage
\onecolumn
\begin{tcolorbox}[
    breakable,         
    title=Negative Constraint Generation,
    colback=gray!5,
    colframe=gray!50,
    fontupper=\small\ttfamily,
]

You are an expert Exam Designer and Logic Analyst. Your goal is to look at the [Question] and predict 1-3 critical pitfalls a planner might make related to risky logical assumptions. \\
Instead of stating obvious generalities, you must identify content-specific risks where a simple keyword search or superficial reading would lead to a wrong answer. \\

Strict Constraints:\\
1. No Hallucination: Do not assume specific text structures (e.g., "split into two").\\
2. No Solutions: Identify the trap only. Do NOT provide actionable plans here.\\
3. No Repetition: Identify assumptions *unique* to this question, not just copying examples.\\

Return the result as a concise JSON list.\\
Format:\\
\{"assumption\_pitfalls": [\\
  "<Pitfall 1: A brief explanation of the pitfall>",\\
  "<Pitfall 2: (Optional)>",\\
  "<Pitfall 3: (Optional)>"\\
]\}\\

-{}-{}- \\
\#\#\# Example 1 (Multiple Inferences Required)\\

[Question]\\
"Why did the author write the article?"\\

[Answer]\\
\{“assumption\_pitfalls": [\\
  "Assuming the author's 'purpose' or 'reason' is stated explicitly as a single sentence.",\\
  "Assuming the 'author' and the 'narrator' are the same entity and share the same motivations.",\\
  "Assuming the purpose must be inferred only from the conclusion, and not from the article's overall tone and main theme."\\
]\}\\

-{}-{}-\\
\#\#\# Example 2 (Specific Scope Constraint)\\

[Question]\\
"What is the “space cafard” that Si describes?"\\

[Answer]\\
\{“assumption\_pitfalls": [\\
  "Assuming any general definition of 'space cafard' is correct, rather than focusing only on the specific definition or description provided by Si."\\
]\}\\

-{}-{}-\\
\#\#\# Example 3 (Multiple Subjects / Scattered Info)\\

[Question]\\
"Arvid 6 and Tendal 13 can perform all of the following abilities EXCEPT:"\\

[Answer]\\
\{“assumption\_pitfalls": [\\
  "Assuming the abilities for 'Arvid 6' and 'Tendal 13' are listed together in the same section.",\\
  "Assuming a single list covers both characters, overlooking the possibility that their details are distributed across different parts of the text."\\
]\}\\

-{}-{}-\\
\#\#\# Example 4 (Evaluation \& Definition)\\

[Question]\\
"Of the following options, which seems to be Tremaine's biggest asset in his investigation?"\\

[Answer]\\
\{“assumption\_pitfalls": [\\
  "Assuming the 'biggest' asset is explicitly labeled as such, rather than needing to first list all assets and then infer their contribution.",\\
  "Assuming 'asset' refers only to physical tools or skills, and overlooking abstract assets like 'intuition', 'connections', or 'reputation'."\\
]\}\\

-{}-{}-\\
\#\#\# Example 5 (Complex Prediction)\\

[Question]\\
"Out of the choices below, predict which future career Eddie would most likely pick given his interests present in the article."\\

[Answer]\\
\{“assumption\_pitfalls": [\\
  "Assuming the prediction relies solely on explicitly stated 'interests', overlooking implied 'skills' or 'aptitudes' mentioned in the text.",\\
  "Assuming the decision is based only on positive factors, failing to account for tasks Eddie explicitly 'dislikes' or 'avoids' which act as negative constraints.",\\
  "Assuming the 'most likely' career is explicitly stated as a future goal, rather than requiring a probabilistic prediction based on his comprehensive profile."\\
]\}\\

-{}-{}-\\
\#\#\# Example 6 (Counting / Manual Aggregation)\\

[Question]\\
"How many times has Critten been a Nilly?"\\

[Answer]\\
\{“assumption\_pitfalls": [\\
  "Assuming the text provides a pre-calculated total count (e.g., 'three times'), rather than requiring the reader to find and count individual mentions manually."\\
]\}\\

-{}-{}-\\
\#\#\# Your Task\\

[Question]\\
\{question\}\\

[Answer]\\

\end{tcolorbox}
\captionof{table}{Detailed prompt for the Pitfall Predictor of \alg.}
\label{tab:neg_const}
\begin{tcolorbox}[
    breakable,         
    title=Constraint-Aware Plan Generation,
    colback=gray!5,
    colframe=gray!50,
    fontupper=\small\ttfamily,
]

[Actions]\\
\{action\_list\}\\
* Note: The output of each action can be the input to other actions.\\

[Instructions]\\
Suppose you are given a question about an article, as well as a list of potential actions (shown above) that you can execute to solve the question. You can imagine the actions as functions in a program, where you have input arguments, as well as output. The output of an action can be fed as input to another action. \\
Please present a sequence of actions that you would use to answer the question after you read the article. The sequence of actions should be specific and cover all the details about the question. Please prioritize using the actions presented in the list above. If you need to add new actions, please follow the format below. Please assign the output of each action with a distinct name, which can be passed into other actions as argument.\\
Think twice before you provide your answer. Make sure your answer is valid, clear, and easy to understand. Keep the answer simple and remove any unnecessary steps. Do not use list comprehension or dictionary comprehension. Keep each action minimally simple. If a question is unanswerable (e.g., requires options), collect as much information as possible from the input such that it will be answerable when provided with options.\\
Since you will be provided with specific pitfalls that must be avoided, please start by performing a brief [Strategy Reasoning]. In this section, explicitly think about how to modify your plan or add specific actions to satisfy these constraints. Once you have established this strategy, proceed to generate the final sequence of actions in the [Answer] section.\\

Your answer should follow the format:\\
\texttt{\textasciigrave\textasciigrave\textasciigrave}\\
{}[Strategy Reasoning]\\
(Briefly explain how you will address the provided pitfalls here)\\

{}[Answer]\\
New actions:\\
- new\_action\_1(arguments) : [one-sentence general explanation] or "-None" if there no need to add new actions\\
- new\_action\_2(arguments) : [one-sentence general explanation] or "-None" if there no need to add new actions\\

1. output\_1 = action\_1(here goes arguments) : [one-sentence explanation]\\
2. output\_2 = action\_2(here goes arguments) : [one-sentence explanation]\\
...\\
\texttt{\textasciigrave\textasciigrave\textasciigrave}\\

The following are a few examples:\\

-{}-{}-\\
Question: "Why is Si retirement so significant to the Space Exploration Team?"\\
Input Pitfalls:\\
- Assuming the significance is stated in a single sentence explicitly linking retirement to the team.\\
- Ignoring the separate chain of events: the reason for retirement and its subsequent consequences.\\

[Strategy Reasoning]\\
The pitfalls warn against looking for a simple, direct link. To address this, I need to split the search: first find the 'cause' of retirement, then find the 'impact' of retirement separately. Then, I must explicitly connect both findings to the 'Space Exploration Team' to synthesize the full answer.\\

[Answer]\\
New actions:\\
- None\\

1. retire\_reason = FIND\_ELEMENT(CTX, "cause", "Si retirement") : Find and summarize the cause or reason of Si retirement from the input article\\
2. retire\_outcome = FIND\_IMPACTS(CTX, "Si retirement") : Find and summarize the impact or outcome or consequences of Si retirement from the input article\\
3. connect\_reason = FIND\_RELATION(CTX, retire\_reason, "Space Exploration Team") : Find and summarize how the reason of Si retirement is related to the Space Exploration Team\\
4. connect\_outcome = FIND\_RELATION(CTX, retire\_outcome, "Space Exploration Team") : Find and summarize how the outcome of Si retirement is related to the Space Exploration Team\\
5. ans = CONCAT(connect\_reason, connect\_outcome) : Combine the previous two steps to form the final answer\\\\

-{}-{}-\\
Question: "What is the “space cafard” that Si describes?"\\
Input Pitfalls:\\
- Assuming any general definition of 'space cafard' is correct.\\
- Failing to restrict the search to only Si's specific description provided in the text.\\

[Strategy Reasoning]\\
The constraint emphasizes avoiding general definitions. Therefore, I must restrict the `FIND\_ELEMENT` action to look specifically for "Si's description" of the term, ensuring the source of the definition is strictly from the character Si in the text.\\

[Answer]\\
New actions:\\
- None\\

1. space\_cafard = FIND\_ELEMENT(CTX, "Si's description", "space cafard") : Find and summarize all relevant information about the "space cafard" strictly as described by Si\\
2. space\_cafard\_cmprh = COMPREHEND(CTX, space\_cafard) : Provide a comprehension about the "space cafard" based on the findings\\
3. ans = CONCAT(space\_cafard, space\_cafard\_cmprh) : Combine to form the final answer\\\\

-{}-{}-\\
Question: "How many times has Critten been a Nilly?"\\
Input Pitfalls:\\
- Assuming the total count (e.g., '3 times') is explicitly stated in the text.\\
- Assuming the plan can just 'search' for a number.\\

{}[Strategy Reasoning]\\
The pitfall indicates that a simple search for a number will fail because the total count isn't explicit. I must first use `FIND\_ALL\_ISSUES` to locate every individual instance where Critten was a Nilly, and then use `COUNT\_X` to manually count these instances.\\

{}[Answer]\\
New actions:\\
- FIND\_ALL\_ISSUES(CTX, X) : Find and summarize all the events where X occurs in the input article\\
- COUNT\_X(CTX, X) : Count the number of times that X occurs in the input article\\\\

1. all\_nilly = FIND\_ALL\_ISSUES(CTX, "Critten been a Nilly") : Find and summarize all individual events/mentions where Critten has been a Nilly\\
2. num\_nilly = COUNT\_X(CTX, all\_nilly) : Count the number of times that Critten has been a Nilly given the collected events above\\\\

-{}-{}-\\
Question: "Out of the choices below, predict which future career Eddie would most likely pick given his interests present in the article."\\
Input Pitfalls:\\
- Assuming only explicitly stated 'interests' matter for the prediction.\\
- Assuming the prediction should be based only on positive factors, ignoring things he dislikes.\\

{}[Strategy Reasoning]\\
The constraints highlight that relying solely on "interests" is insufficient. I must modify the plan to actively search for "skills/aptitudes" (implied interests) and "dislikes/avoids" (negative constraints). These additional factors must be concatenated into the profile before making a prediction.\\

{}[Answer]\\
New actions:\\
- PREDICT\_CAREER(CTX, X, Y) : Predict the future career given a person X's future career according to their interests or goals Y\\

1. eddie = IDENTIFY\_ELEMENT(CTX, "Eddie") : Identify who Eddie is in the input article\\
2. eddie\_interests = FIND\_ELEMENT(CTX, "interests", eddie) : Find and summarize all the interests of Eddie\\
3. eddie\_skills = FIND\_ELEMENT(CTX, "skills and aptitudes", eddie) : Find demonstrated skills or aptitudes, as required to avoid the pitfall of missing implied traits\\
4. eddie\_dislikes = FIND\_ELEMENT(CTX, "dislikes and avoids", eddie) : Find tasks Eddie dislikes, as required to filter out unlikely careers\\
5. eddie\_goals = FIND\_INTENT(CTX, eddie) : Find and summarize the intent/purpose/goal of Eddie\\
6. eddie\_profile = CONCAT(eddie\_interests, eddie\_skills, eddie\_dislikes, eddie\_goals) : Combine interests, skills, dislikes, and goals to build a complete profile\\
7. ans = PREDICT\_CAREER(CTX, "Eddie", eddie\_profile) : Predict the future career based on the comprehensive profile\\\\

-{}-{}-\\
Question: "Which word doesn't describe the security guard?"\\
Input Pitfalls:\\
- Assuming the plan should search for words that *do not* describe the guard directly.\\
- Failing to understand this is a 'NOT' (exclusion) question requiring a list of valid descriptions first.\\\
{}[Strategy Reasoning]\\
The pitfall warns against searching for the negative directly. The correct strategy is to first find all words that *DO* describe the security guard in the text. Then, the final answer (likely comparing with options later) will be derived from knowing what *is* true.\\

{}[Answer]\\
New actions:\\
- None\\

1. security\_guard = FIND\_CHARACTER(CTX, "security guard") : Find and summarize the character traits of the security guard\\
2. guard\_descriptions = FIND(CTX, "descriptive words", "security guard") : Find the words that ARE used to describe the security guard in the text\\
3. ans = CONCAT(security\_guard, guard\_descriptions) : Combine the traits and descriptions to form a basis for exclusion\\\\

-{}-{}-\\
Question: "Of the following options, which seems to be Tremaine's biggest asset in his investigation?"\\
Input Pitfalls:\\
- Assuming 'asset' refers only to physical tools.\\
- Assuming the 'biggest' asset is explicitly labeled as such.\\

{}[Strategy Reasoning]\\
To avoid the pitfall of focusing only on physical tools, I must explicitly instruct the `FIND\_ELEMENT` action to look for "assets including abstract ones (intuition, connections)". Also, since the "biggest" isn't labeled, I need to `SORT` the found assets based on their impact to determine the ranking.\\

{}[Answer]\\
New actions:\\
- SORT(CTX, X): Sort the elements in X in ascending order with concise reasons, based on the input article\\

1. tremaine = IDENTIFY\_ELEMENT(CTX, "Tremaine") : Identify who Tremaine is in the input article\\
2. tremaine\_assets = FIND\_ELEMENT(CTX, "assets (physical and abstract)", tremaine) : Find all assets, explicitly including abstract ones like intuition or connections\\
3. ranked\_assets = SORT(CTX, tremaine\_assets) : Sort the assets in ascending order of importance/impact based on the text\\\\

{}[Question]\\
Now you are given a question about an article:\\
    \{question\}\\
You MUST avoid these core pitfalls identified for this question:\\
\{assumption\_pitfall\}\\

Please provide a plan (sequence of actions) that can arrive to the answer after reading the article. Before generating the final plan, please briefly analyze how to address these pitfalls in a [Strategy Reasoning] section. Then, provide the final sequence in the [Answer] section. As the corresponding options are not provided for the question, when the question is not answerable without the options, simply collect as much information as possible from the input such that it will be answerable with the options. Make sure the plan you generate is valid and faithful to the question.
\\\\
{}[Strategy Reasoning]\\

\end{tcolorbox}
\captionof{table}{Detailed prompt for the Planner of \alg.}
\label{tab:CA-Plan}
\begin{tcolorbox}[
    breakable,         
    title=Context-Aware Correction,
    colback=gray!5,
    colframe=gray!50,
    fontupper=\small\ttfamily,
]

[Actions]\\
\{action\_list\}\\
* Note: The output of each action can be the input to other actions.\\

{}[Instructions]\\
Suppose you are given a question about an article, as well as a list of potential actions (shown above) that you can execute to solve the question. You can imagine the actions as functions in a program, where you have input arguments, as well as output. The output of an action can be fed as input to another action.\\
Please present a sequence of actions that you would use to answer the question after you read the article. The sequence of actions should be specific and cover all the details about the question. Please prioritize using the actions presented in the list above. If you need to add new actions, please follow the format below. Please assign the output of each action with a distinct name, which can be passed into other actions as argument.
Think twice before you provide your answer. Make sure your answer is valid, clear, and easy to understand. Keep the answer simple and remove any unnecessary steps. Do not use list comprehension or dictionary comprehension. Keep each action minimally simple. If a question is unanswerable (e.g., requires options), collect as much information as possible from the input such that it will be answerable when provided with options.\\
Since you will be provided with an invalid plan and parser error messages, please start by performing a brief [Strategy Reasoning]. In this section, identify the cause of the error and plan how to fix the syntax while preserving the original logic. Once you have established this repair plan, proceed to generate the corrected sequence of actions in the [Answer] section.\\

Your answer should follow the format:\\
\texttt{\textasciigrave\textasciigrave\textasciigrave}\\
{}[Strategy Reasoning]\\
(Briefly explain how you will address the provided pitfalls here)\\

{}[Answer]\\
New actions:\\
- new\_action\_1(arguments) : [one-sentence general explanation] or "-None" if there no need to add new actions\\
- new\_action\_2(arguments) : [one-sentence general explanation] or "-None" if there no need to add new actions\\

1. output\_1 = action\_1(here goes arguments) : [one-sentence explanation]\\
2. output\_2 = action\_2(here goes arguments) : [one-sentence explanation]\\
...\\
\texttt{\textasciigrave\textasciigrave\textasciigrave}\\

The following are examples of how to correct an invalid plan based on error messages:\\

-{}-{}-\\
\#\#\# Example 1 (Error: Unknown Action)\\
Question: "What is the primary diet of the spectacled bear?"\\

Invalid Plan:\\
1. bear\_info = FIND\_ELEMENT(CTX, "diet", "spectacled bear") : Find diet info\\
2. ans = COMPREHEND(CTX, bear\_info) : Understand the info\\

Error Message:\\
"Error parsing action COMPREHEND. Unknown action. Please define it in the 'New actions' section if needed, or choose from the existing action list."\\

Input Pitfalls:\\
- Assuming the diet consists of only one type of food.\\

{}[Strategy Reasoning]\\
The parser reports that `COMPREHEND` is an unknown action. Checking the valid action list, the correct action for summarizing information is `SUMMARIZE`. I will replace `COMPREHEND` with `SUMMARIZE` while keeping the arguments the same to preserve the flow.\\

{}[Answer]\\
New actions:\\
- None\\

1. bear\_info = FIND\_ELEMENT(CTX, "diet", "spectacled bear") : Find diet info\\
2. ans = SUMMARIZE(CTX, bear\_info) : Summarize the findings to form the answer\\\\

-{}-{}-\\
\#\#\# Example 2 (Error: Undefined Variable)\\
Question: "How did the protagonist escape the room?"\\

Invalid Plan: \\
1. room\_desc = FIND\_ELEMENT(CTX, "escape method", "protagonist") : Find escape details\\
2. ans = GENERATE\_ANSWER(CTX, room\_info) : Generate the final answer\\

Error Messages:\\
"Error parsing action GENERATE\_ANSWER. Argument room\_info is not defined."\\

Input Pitfalls: \\
- Assuming the escape happened in a single step.\\

{}[Strategy Reasoning]\\
The error states that `room\_info` is undefined. Looking at the previous step (step 1), the output variable was named `room\_desc`, not `room\_info`. I must correct the argument in step 2 to match the variable name defined in step 1 (`room\_desc`).\\

{}[Answer]\\
New actions:\\
- None\\

1. room\_desc = FIND\_ELEMENT(CTX, "escape method", "protagonist") : Find escape details\\
2. ans = GENERATE\_ANSWER(CTX, room\_desc) : Generate the final answer\\\\

-{}-{}-\\
\#\#\# Example 3 (Error: Incorrect Argument Count)\\
Question: "List all the awards won by the author."\\

Invalid Plan: \\
1. awards = FIND\_ALL\_ISSUES("awards", "author") : Find all awards\\
2. ans = LIST\_ITEMS(CTX, awards) : List them\\

Error Message:\\
"Error parsing action FIND\_ALL\_ISSUES. Number of arguments is incorrect"\\

Input Pitfalls:\\
- Assuming the awards are listed in a distinct 'awards' section.\\

{}[Strategy Reasoning]\\
The action `FIND\_ALL\_ISSUES` caused an argument count error. Standard actions usually require `CTX` as the first argument. I will add `CTX` to the arguments of `FIND\_ALL\_ISSUES` to satisfy the function definition.\\

{}[Answer]\\
New actions:\\
- None\\

1. awards = FIND\_ALL\_ISSUES(CTX, "awards", "author") : Find all awards\\
2. ans = LIST\_ITEMS(CTX, awards) : List them\\\\

-{}-{}-\\
\#\#\# Example 4 (Error: Missing Action Definition)\\
Question: "Based on the historical data provided, predict the stock price for next month."\\

Invalid Plan:\\
1. history = FIND\_DATA(CTX, "stock price history", "last 5 years") : Retrieve data\\
2. prediction = PREDICT\_TREND(CTX, history) : Predict future price\\
3. ans = GENERATE\_ANSWER(CTX, prediction) : Formulate answer\\

Error Message:\\
"Error parsing action PREDICT\_TREND. Unknown action. Please define it in the 'New actions' section if needed, or choose from the existing action list."\\

Input Pitfalls:\\
"Assuming a linear trend without considering volatility mentioned in the text."\\

{}[Strategy Reasoning]\\
The parser indicates PREDICT\_TREND is unknown. This action is essential for the prediction task and has no equivalent in the standard action list. Therefore, instead of replacing it, I will retain PREDICT\_TREND and explicitly define it in the 'New actions' section to fix the error.\\

{}[Answer]\\
New actions:\\
- PREDICT\_TREND(CTX, data) : Analyze the provided data to forecast future trends or values.\\

1. history = FIND\_DATA(CTX, "stock price history", "last 5 years") : Retrieve data\\
2. prediction = PREDICT\_TREND(CTX, history) : Predict future price based on the retrieved history\\
3. ans = GENERATE\_ANSWER(CTX, prediction) : Formulate the final answer\\\\

{}[Question]\\
Given the following question, \\
Question: \{question\}\\
you just came up with the following sequence of actions as well as potential new actions:\\
\{invalid\_plan\}\\

However, the above answer is invalid according to a parser, which returned an error message:\\
    \{error\_message\}\\

You MUST avoid these core pitfalls identified for this question:\\
\{assumption\_pitfall\}\\

Can you edit the plan (sequence of actions) to make it valid? Please focus strictly on correcting the syntax errors reported by the parser, such as missing arguments, missing descriptions, or incorrect formatting. Do not attempt to generate a new strategy or change the sequence of actions unless it is absolutely required to fix the syntax error (e.g., adding a missing step to define an undefined variable). Your goal is to simply repair the invalid plan so that it becomes parseable while maintaining its original logic.\\

Before generating the final corrected plan, please use the [Strategy Reasoning] section to briefly identify the specific syntax error from the error message and explain how you will correct the format (e.g., adding missing arguments or fixing the action definition) to satisfy the parser. Then, provide the final corrected sequence in the [Answer] section.\\

{}[Strategy Reasoning]\\

\end{tcolorbox}
\captionof{table}{Detailed prompt for the Corrector of \alg.}
\label{tab:CA-Corr}
\clearpage
\begin{tcolorbox}[
    breakable,         
    title=Generative Question Answering (GQA),
    colback=gray!5,
    colframe=gray!50,
    fontupper=\small\ttfamily,
]

Article\\

\{article\}\\

End of Article\\\\

Question: \{question\}\\

Answer:\\
(Please provide a detailed explanation for answering the question above.) \\

\end{tcolorbox}
\captionof{table}{Detailed prompt for Generative Question Answering (GQA).}
\label{tab:GenerativeQA}
\begin{tcolorbox}[
    breakable,         
    title=Chain-of-Thought (CoT),
    colback=gray!5,
    colframe=gray!50,
    fontupper=\small\ttfamily,
]

Article:\\
\{article\}\\

Question: \{question\}\\

Please think step by step to find the answer based on the article. \\
Provide your reasoning process first, and then give the final answer.\\

Reasoning: \\

\end{tcolorbox}
\captionof{table}{Detailed prompt for Chain-of-Thought (CoT).}
\label{tab:CoT}
\begin{tcolorbox}[
    breakable,         
    title=Plan-and-Solve,
    colback=gray!5,
    colframe=gray!50,
    fontupper=\small\ttfamily,
]

Article:\\
\{article\}\\

Question: \{question\}\\

Let’s first understand the problem and devise a plan to solve it. Then, let’s carry out the plan and solve the problem step by step. Please respond in the following format:\\
Plan: [Your plan here]\\
Solution: [Your step-by-step execution here]\\

Answer:\\

\end{tcolorbox}
\captionof{table}{Detailed prompt for Plan-and-Solve.}
\label{tab:Plan-and-Solve}
\begin{tcolorbox}[
    breakable,         
    title=ReAct (action generation),
    colback=gray!5,
    colframe=gray!50,
    fontupper=\small\ttfamily,
]

{}[Actions]\\
\{action\_list\}\\
FINISH() : Returns the answer and finishes the task.\\
* Note: The output of each action can be the input to other actions.\\

{}[Instructions]\\
Solve the question by interleaving Thought, Action, and Observation steps.\\
- Thought i: Reason about the current situation and decide what to do next.\\
- Action i: Choose an action from the available toolset to gather information. \\
- Observation i: The output or result from the environment based on your action.\\

{}[Format Rules]\\
- Continue from `Thought \{step\_idx\}:`.\\
- Output exactly one `Thought \{step\_idx\}: ...` line and one `Action \{step\_idx\}: ...` line.\\
- Do not generate `Observation \{step\_idx\}:` yourself.\\
- To finish, use `Action \{step\_idx\}: FINISH(your final answer)`.\\

The following are a few examples\\

-{}-{}-\\
Question: "Why is Si retirement so significant to the Space Exploration Team?"\\

Thought 1: I need to find the cause or reason for Si's retirement from the article first.\\
Action 1: FIND\_ELEMENT(CTX, "cause", "Si retirement")\\
Observation 1: Si retired due to health issues after 30 years of service, which was unexpected for the team.\\
Thought 2: Now that I know the reason, I should find out the impact or consequences of this retirement to understand its significance.\\
Action 2: FIND\_IMPACTS(CTX, "Si retirement")\\
Observation 2: The team lost its most experienced navigator, leading to a delay in the Mars mission.\\
Thought 3: I have the reason and the outcome. Now I need to connect the reason specifically to the Space Exploration Team.\\
Action 3: FIND\_RELATION(CTX, "health issues", "Space Exploration Team")\\
Observation 3: The team had no backup plan for such a sudden medical retirement of a key member.\\
Thought 4: I have collected all the reasons and their relations. I can now provide the final answer.\\
Action 4: FINISH()\\

-{}-{}-\\
Question: "What is the “space cafard” that Si describes?"\\

Thought 1: I need to find all relevant information and Si's specific description of "space cafard" from the article.\\
Action 1: FIND\_ELEMENT(CTX, "Si's description", "space cafard")\\
Observation 1: Si describes "space cafard" as a peculiar mental strain or madness that affects astronauts during long periods of isolation in deep space.\\
Thought 2: Now that I have the basic description, I should provide a deeper comprehension of this concept based on the input to fully answer the question.\\
Action 2: COMPREHEND(CTX, "mental strain and madness during isolation")\\
Observation 2: It refers to the psychological breakdown where the vastness of space becomes overwhelming, leading to erratic behavior and a loss of reality.\\
Thought 3: I have gathered both the direct description and the deeper comprehension. I can now provide the final answer by combining these points.\\
Action 3: FINISH()\\

-{}-{}-\\
Question: "How does Johnson trick the duo into paying for things more than once?"\\

Thought 1: First, I need to identify and summarize the specific events where Johnson tricks the duo.\\
Action 1: FIND\_EVENT(CTX, "Johnson tricks the duo")\\
Observation 1: Johnson used falsified invoices and claimed there were "administrative errors" that required immediate re-payment for already settled services.\\
Thought 2: Next, I need to find the events where the duo actually paid for things to see the instances of double payment.\\
Action 2: FIND\_EVENT(CTX, "the duo pay for things")\\
Observation 2: The duo paid for their docking fees and fuel supplies twice, once to the automated system and once directly to Johnson.\\
Thought 3: Now I should analyze the connection/relation between Johnson's trickery and the duo's payments to understand the mechanism.\\
Action 3: FIND\_RELATION(CTX, "falsified invoices", "paying docking and fuel fees twice")\\
Observation 3: Johnson intercepted the digital receipts and convinced the duo that their first payments had failed to process.\\
Thought 4: I will now summarize the specific trick method used by Johnson in these events to get a detailed explanation.\\
Action 4: FIND\_ELEMENT(CTX, "intercepting receipts and claiming failure", "trick method")
Observation 4: The trick method involved exploiting the duo's lack of familiarity with the station's new payment interface and creating a false sense of urgency.\\
Thought 5: I have all the components: the tricking events, the payment events, their relation, and the specific method. I am ready to form the final answer.\\
Action 5: FINISH()\\

-{}-{}-\\
Question: "Why did the author write the article?"\\

Thought 1: To understand the author's purpose, I first need to identify the central theme of the article.\\
Action 1: IDENTIFY\_ELEMENT(CTX, "theme")\\
Observation 1: The theme is the psychological impact of deep-space isolation on human relationships.
Thought 2: Now that I know the theme, I should find the moral or the intended lesson the author wants to convey.\\
Action 2: FIND\_MORAL(CTX)\\
Observation 2: The lesson is that maintaining human connection is vital even in the most desolate environments.\\
Thought 3: Next, identifying the author's tone will help clarify the intent behind the writing.
Action 3: IDENTIFY\_ELEMENT(CTX, "tone")\\
Observation 3: The tone is introspective and cautionary, emphasizing the fragility of the mind.
Thought 4: Finally, I need to identify the target audience to fully grasp why the author chose this specific message.\\
Action 4: IDENTIFY\_ELEMENT(CTX, "target audience")\\
Observation 4: The article seems to be targeted toward aspiring space travelers and psychologists.
Thought 5: I have identified the theme, moral, tone, and target audience. I can now combine these elements to explain the author's purpose.\\
Action 5: FINISH()\\

-{}-{}-\\
Question: "How many times has Critten been a Nilly?"\\

Thought 1: I need to find and summarize all specific instances or events in the article where Critten is mentioned as being a Nilly.\\
Action 1: FIND\_ALL\_ISSUES(CTX, "Critten been a Nilly")\\
Observation 1: The article describes three separate occasions: first at the ceremony, second during the mission, and finally at the transition.\\
Thought 2: Now that I have summarized the specific events, I will count the number of times these occurrences happened to answer the question.\\
Action 2: COUNT\_X(CTX, "three separate occasions of Critten being a Nilly")\\
Observation 2: The total count of these events is 3.\\
Thought 3: I have found all events and successfully counted them. I am ready to provide the final answer.\\
Action 3: FINISH()\\

-{}-{}-\\
Question: "Arvid 6 and Tendal 13 can perform all of the following abilities EXCEPT:"\\

Thought 1: First, I need to identify what "Arvid 6" refers to in the article to understand its context.\\
Action 1: IDENTIFY\_ELEMENT(CTX, "Arvid 6")\\
Observation 1: Arvid 6 is an advanced surveillance drone designed for high-altitude reconnaissance.\\
Thought 2: Now I should also identify "Tendal 13" to see how it relates to Arvid 6.\\
Action 2: IDENTIFY\_ELEMENT(CTX, "Tendal 13")\\
Observation 2: Tendal 13 is a ground-based support unit that coordinates with Arvid 6 drones.\\
Thought 3: Since I know both entities, I will now find and summarize all the specific abilities of Arvid 6.\\
Action 3: FIND\_ELEMENT(CTX, "abilities", "Arvid 6 surveillance drone")\\
Observation 3: Arvid 6 can perform thermal imaging, long-range signal interception, and autonomous flight.\\
Thought 4: Next, I need to find the abilities of Tendal 13 to have a complete list for comparison.\\
Action 4: FIND\_ELEMENT(CTX, "abilities", "Tendal 13 support unit")\\
Observation 4: Tendal 13 can perform data encryption, remote recharging, and heavy terrain navigation.\\
Thought 5: I have collected all abilities for both Arvid 6 and Tendal 13. I can now compare these with the options to find the exception.\\
Action 5: FINISH()\\

-{}-{}-\\
Question: "The purpose for the narrator losing his memory is. . ."\\

Thought 1: To understand the memory loss, I must first identify who the narrator is and their role in the story.\\
Action 1: IDENTIFY\_CHARACTER(CTX, "narrator")\\
Observation 1: The narrator is a former intelligence officer named Elias who is currently in hiding.
Thought 2: Now I need to find and summarize the specific events where Elias (the narrator) loses his memory.\\
Action 2: FIND\_ELEMENT(CTX, "memory loss", "Elias")\\
Observation 2: Elias undergoes a series of neuro-suppression sessions that gradually erase his recollections of the "Blackwood" operation.\\
Thought 3: I have identified the events. Now I should find the specific purpose or reason behind why the narrator had to lose his memory.\\
Action 3: FIND\_ELEMENT(CTX, "purpose", "neuro-suppression sessions and memory loss")\\
Observation 3: The purpose was to protect classified information from being extracted by enemy interrogators if Elias were captured.\\
Thought 4: I have found the narrator, the memory loss events, and the underlying purpose. I am ready to provide the final answer.\\
Action 4: FINISH()\\\\

{}[Question]\\
Now you are given a question about an article:\\
    Question: \{question\}\\

Solve the task by interleaving Thought, Action, and Observation steps. \\
1. For each step, provide a 'Thought' to reason about the current state, followed by an 'Action' to call a specific function.\\
2. After the 'Action', wait for the 'Observation' (the result of your action). \\
3. Do NOT generate the Observation yourself. \\
4. If the corresponding options are not provided, collect all relevant information until you can definitively answer the question.\\
5. Once you have enough information, use "Action: Finish[answer]" to provide the final result.\\

Begin the process now.\\

{}[Answer]\\
\{last\_response\}\\

\end{tcolorbox}
\captionof{table}{Detailed prompt for action generation of ReAct.}
\label{tab:react_action}
\begin{tcolorbox}[
    breakable,         
    title=ReAct (action generation),
    colback=gray!5,
    colframe=gray!50,
    fontupper=\small\ttfamily,
]

{}[Article]\\
\{article\}\\

{}[Context]\\
- Question: \{question\}\\
- Action to Execute: '\{last\_action\} : \{thought\}'\\

{}[Goal]\\
Based on the Article, provide the specific information requested by the Action.\\
Use the Current Thought only as guidance for what to look for; do not treat it as evidence.\\
Do not add any reasoning or external knowledge. If not found, say "Information not found."\\

Result:

\end{tcolorbox}
\captionof{table}{Detailed prompt for action execution of ReAct.}
\label{tab:react_execution}
\begin{tcolorbox}[
    breakable,         
    title=LLM-as-a-Judge,
    colback=gray!5,
    colframe=gray!50,
    fontupper=\small\ttfamily,
]
Relevant information for answering the question:\\

\{open\_answer\}\\

Question:\{question\}\\
\{options\}\\

Carefully compare all four options (A, B, C, and D) based on the relevant information. Select the best possible answer by ensuring it is the most accurate choice given the information provided. Write only the letter of your final answer without explanation. Answer (select from A, B, C, D):\\

\end{tcolorbox}
\captionof{table}{Detailed prompt for LLM-as-a-Judge in multiple-choice questions.}
\label{tab:LLM-as-a-Judge}
\clearpage

\end{document}